\documentclass[runningheads]{llncs}
\usepackage[T1]{fontenc}
\usepackage{graphicx}
\usepackage{svg}
\usepackage{microtype}
\usepackage{url,hyperref,lineno,adjustbox, multirow, booktabs}
\usepackage[table]{xcolor} 
\usepackage{fix-cm}
\usepackage{amsmath}
\usepackage{amssymb}
\usepackage{esvect} 
\usepackage{subcaption}
\usepackage{float} 
\usepackage{caption}
\usepackage{tabularx}  
\usepackage{wrapfig}

\begin{document}
\title{Unconventional Hexacopters via Evolution and Learning: Performance Gains and New Insights}
\titlerunning{Unconventional Hexacopters via Evolution and Learning}
%
\author{Jed R. Muff\inst{1}\orcidID{0009-0004-9934-6403} \and
Keiichi Ito\inst{1}\orcidID{0000-0003-1364-9477} \and
Elijah H. W. Ang\inst{2}\orcidID{0000-0003-4534-7924} \and
Karine Miras\inst{1}\orcidID{0000-0003-4942-3488} \and
A.E. Eiben\inst{1}\orcidID{0000-0002-3106-4213}}

\authorrunning{J.R Muff et al.}
%
\institute{Department of Computer Science, Vrije Universiteit, Amsterdam, The Netherlands \and
Department of Control and Operations, Technische Universiteit Delft, The Netherlands}

\maketitle              
\begin{abstract}
This study investigates a system of hexacopter type drones with evolvable morphologies and learnable controllers. The combination of morphological evolution and reinforcement learning is shown to produce unconventional drones that significantly outperform the traditional hexacopter on several tasks that are more complex than previously considered in the literature. In addition, novel metrics are introduced and new analyses are conducted on the interaction between morphological evolution and learning, uncovering previously unidentified effects. 

\end{abstract}

\section{Introduction}

The motivation behind this study lies in the need for high performing drone designs tailored to diverse navigation tasks. Given the achievements of evolutionary robotics (ER), it is logical to try to use evolutionary algorithms to optimize drone designs; however, the ER research community has a dominant focus on land robots. From this perspective, this paper addresses a new application domain for ER, exploring the potential of evolutionary methods for aerial robots.  

Our main hypothesis is that \textit{evolution can produce unconventional drone morphologies that outperform the traditional hexacopter design}. To this end, we adopt an evolutionary robotics framework that combines evolution and learning~\cite{Eiben2013,Gupta2021} to discover novel and high performing morphologies that conventional design principles might overlook. Performance is evaluated on four navigation tasks, each defined by a racetrack with a different geometric structure: figure-eight, circle, shuttle run, and slalom. 
The first objective of this paper is to test whether evolution can indeed discover unconventional morphologies that outperform traditional designs across these navigation tasks.

The second objective is to gain insights into the combination of morphological evolution and learning. While the relationship between evolution and learning has been widely studied, a concrete understanding remains elusive~\cite{Sznajder_Sabelis_Egas_2011,Whiten2007}. This may be partially caused by the complex dual role morphology plays in such a system: It not only influences performance directly but also indirectly, by affecting the learnability of effective controllers. The concept of morphological intelligence highlights that certain body configurations can simplify control or embed task relevant dynamics into the physical structure of the system~\cite{pfeifer2006body,pfeifer2006morphological}. 

As for the learning component, note that most existing systems either evolve morphology and control jointly without lifetime learning, or they add learning of controllers to systems where morphology and control co-evolve. To mitigate the resulting complexity, our approach instead algorithmically decouples the evolutionary search for body designs from the learning of effective control systems. Specifically,
we apply evolution exclusively to the morphologies, and use reinforcement learning (from scratch) to obtain a good controller for each `newborn' drone body. This division of adaptation mechanisms provides two benefits: 1) allowing to isolate the role of the body in determining fitness and 2) simplifying the study how evolution and learning interact. 

The main contribution of this work is to demonstrate how evolution can be used to improve drone morphology design, while helping to clarify the principles through which such improvement occurs. Although the current simulation based results still need to be validated in real hardware, which is beyond the scope of this study, these results represent a crucial step toward applying evolutionary design in a new domain of areal robotics.

\section{Related Work}
\label{sec:SOA}
We are not to compare evolutionary and traditional hexacopter design {\it methods} directly. To assess our main hypothesis, it is sufficient to compare {\it outcomes}: the evolved drones with a classic hexacopter, allowing them to use the same learning algorithm with the same learning budget to adapt to each of the tasks separately.

The research area relevant for our second objective is the combination of morphological evolution and learning. This is a classic topic in natural and artificial systems~\cite{1987-When-learning-guides-evolution,Hinton1987,1999-learning-and,2003-Evolution-and-Learning}. Although learning was long seen as an optional feature in artificial evolution, with both advantages and disadvantages~\cite{1996-Maturaion,1997-Landscapes}, its importance became clear when evolutionary robotics moved beyond evolving controllers for fixed morphologies~\cite{Floreano1996,prabhu2018survey} toward the simultaneous evolution of morphology and control, where learning emerged as a crucial ingredient: `{\it If it evolves, it needs to learn}'~\cite{EibenHart2020}. This shift highlighted the deep connection between a robot’s body and its ability to learn, what has been termed morphological intelligence. Subsequent studies confirmed that combining evolution with learning not only increases fitness but also produces qualitatively different morphologies~\cite{Miras2020EvoLearn,JLO-2022,Cheney2018}. Gupta et al.~\cite{Gupta2021} demonstrated the Baldwin effect, where morphologies evolved to make learning new tasks faster over generations. They defined morphological intelligence by the ability to learn new tasks, while the so-called learning delta introduced by Miras et al.~\cite{Miras2020EvoLearn} related it to the given task for which robots are being evolved. These studies establish a key principle: evolution can discover morphologies that are inherently easier to control, effectively shaping the learning landscape itself (or equally, learning shaping the course of evolution)~\cite{Hinton1987}.

\section{Problem Formulation and Phenotype Space}
Our design problem can be naturally decomposed into two interacting components: the \textbf{morphology} (the physical body of the drone, including the central unit, arms, motors, and propellers) and the \textbf{controller} (the software or ``brain'' that generates motor commands). The performance of a drone emerges from the interplay between these two components, which makes effective design dependent on their joint optimization. In this work, we adopt complementary approaches: morphologies are evolved, while controllers are not evolved but learned through reinforcement learning. 

A drone’s morphology can be regarded as one part of the phenotype, with the primary variable component being the placement of its motors. Each motor–propeller unit is defined by a three-dimensional position and orientation relative to the central core. While the core remains constant across all morphologies, the arms and motor configurations vary. To specify an evolutionary algorithm for this setting, it is necessary to define suitable genotypes, as well as reproduction operators to generate new genotypes from existing ones, and selection operators to regulate survival and reproduction. The details of these elements, together with the overall fitness function, are provided in section~\ref{sec:evo}.

In parallel, the controller of the drone constitutes the second component of the phenotype. The controller, or policy network in reinforcement learning terminology, is implemented as a three-layer multilayer perceptron. It consists of an input layer with 16 neurons corresponding to the observation space (described below), three fully connected hidden layers of 64 neurons each with ReLU activations, and an output layer of six neurons (one per motor command) with linear activation. The motor commands are subsequently scaled to the [0,1] range to match the normalized propeller speed requirements.

\begin{figure}[ht]
 \centering
 \includegraphics[width=.6\linewidth]{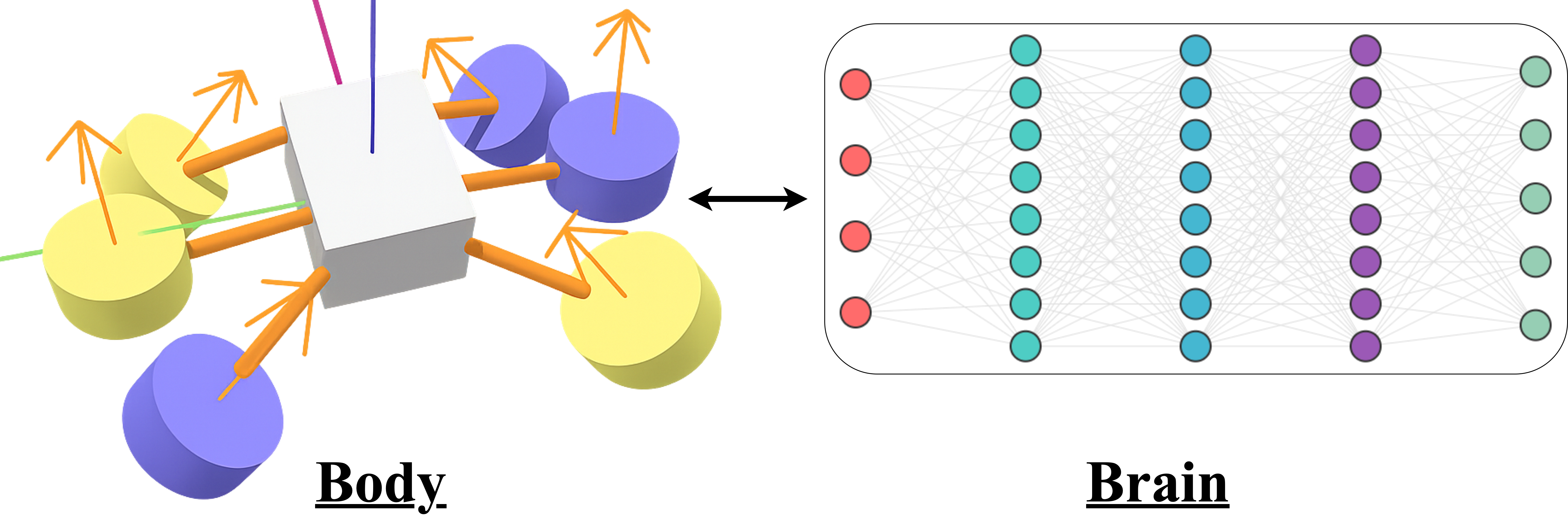}
 \caption{Hexacopter type drone body and multilayer perceptron-type brain.}
 \label{fig:phenotypes}
\end{figure}

The tunable parameters of this network include the weight matrices $W_1 \in \mathbb{R}^{16 \times 64}$, $W_2 \in \mathbb{R}^{64 \times 64}$, $W_3 \in \mathbb{R}^{64 \times 64}$, and $W_4 \in \mathbb{R}^{64 \times 6}$, together with their associated bias vectors $b_1 \in \mathbb{R}^{64}$, $b_2 \in \mathbb{R}^{64}$, $b_3 \in \mathbb{R}^{64}$, and $b_4 \in \mathbb{R}^{6}$. This architecture results in a total of $(16 \times 64) + 64 + (64 \times 64) + 64 + (64 \times 64) + 64 + (64 \times 6) + 6 = 9,798$ trainable parameters that are optimized during the RL process. The optimization of the policy will be described in section~\ref{sec:rl}.

\begin{figure}[hbt!]
\centering
\begin{minipage}[b]{0.15\textwidth}
    \centering
    \includegraphics[width=\linewidth]{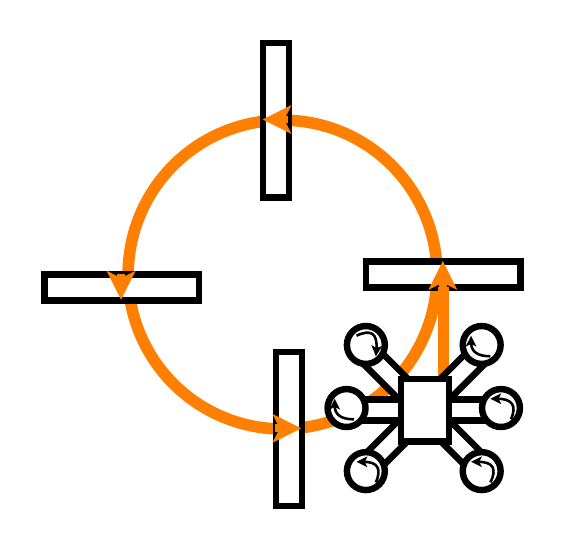}
    \textbf{(a)} Circle task.
    \label{circle_task_drawing}
\end{minipage}
\begin{minipage}[b]{0.35\textwidth}
    \centering
    \includegraphics[width=\linewidth]{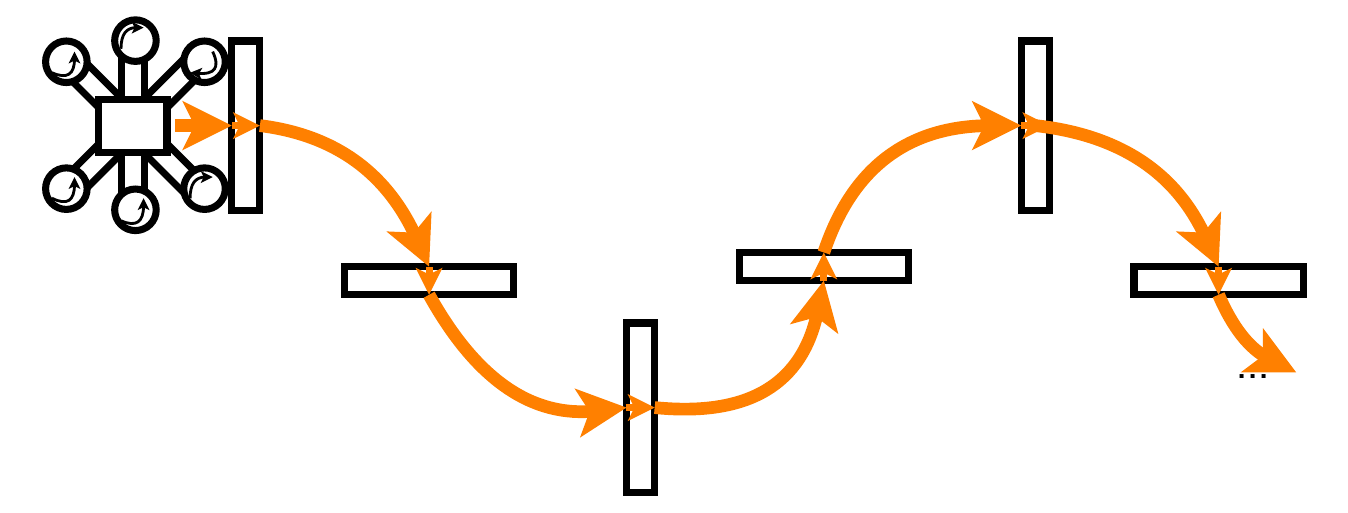}
    \refstepcounter{subfigure}
    \textbf{(b)} Slalom task.
    \label{slalom_task_drawing}
\end{minipage}
\begin{minipage}[b]{0.2\textwidth}
    \centering
    \includegraphics[width=\linewidth]{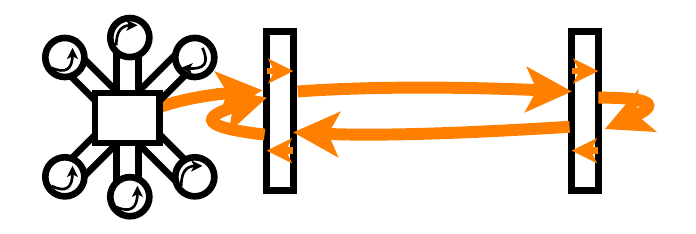}
    \refstepcounter{subfigure}
    \textbf{(c)} Shuttlerun task
    \label{shuttlerun_task_drawing}
\end{minipage}
\begin{minipage}[b]{0.25\textwidth}
    \centering
    \includegraphics[width=\linewidth]{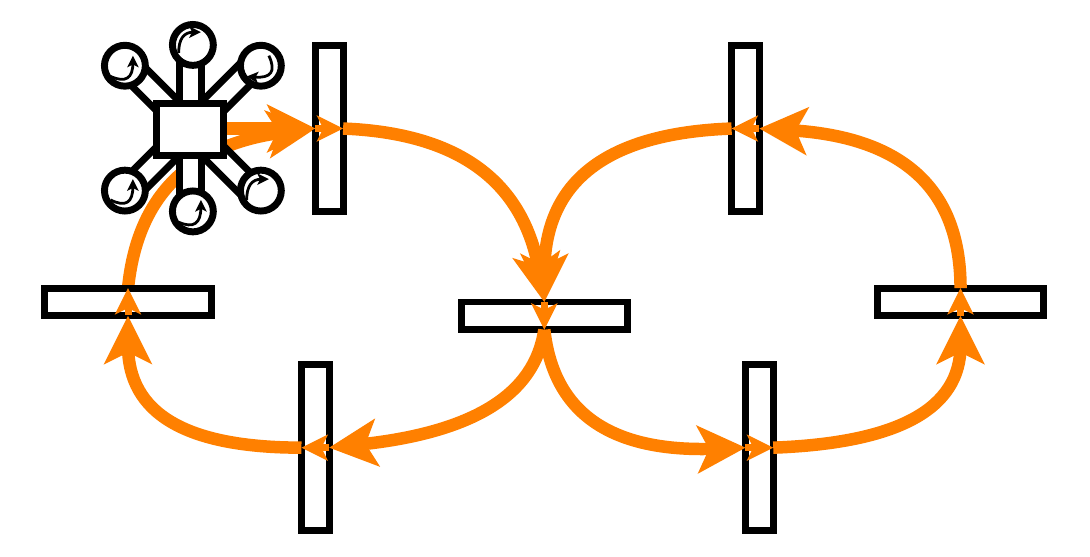}
    \refstepcounter{subfigure}
    \textbf{(d)} Figure8 task.
    \label{figure8_task_drawing}
\end{minipage}

\caption{Diagrammatic representation of the tasks.}
\label{task_drawings}
\end{figure}

We consider four tasks shown in figure \ref{task_drawings}: a circular track where the goal is to travel in a circle, a slalom track where the goal is to weave in and out along a line like a snake, a shuttlerun task where the drone has to fly back and forth between two gates, and an infinity track or figure8 track.

These four tasks were chosen as fundamental benchmarks for assessing drone navigation capabilities. Although relatively simple and structured, each task targets essential flight skills that underpin more advanced behaviors: the circular flight task evaluates sustained asymmetric curved trajectories; the slalom task assesses agility and rapid directional changes; the shuttlerun measures acceleration, deceleration, and sharp turning; and the figure-eight task integrates multiple directional shifts for a well rounded challenge. The regular and predictable nature of these patterns facilitates clear performance comparisons between conventional and evolved drone designs, serving as a proof of concept before transitioning to more complex and irregular flight scenarios. In addition, these tasks enable an interpretable analysis of how specific morphological traits confer advantages, insights that would be harder to extract in more chaotic or unstructured environments.

\section{Evolutionary Process}
\label{sec:evo}

To evolve drone morphologies, we employ a $(\mu + \lambda)$ evolution strategy where parent selection is performed through tournament selection with a tournament size of 3. We incorporate learning following the Triangle of Life model \cite{Eiben2013}; this includes a learning phase where each morphology develops appropriate control behaviors before fitness evaluation. Fitness is quantified as the number of waypoints successfully traversed during a standardized 12-second evaluation period. This task agnostic metric provides a generalizable framework applicable to any waypoint based navigation scenario, enabling consistent evaluation across diverse flight tasks and environmental conditions. The remaining evolutionary components are detailed in the following. 

\subsubsection{Representation and Reproduction}
\label{sec:repr}
We represent the morphology of a drone as a two-dimensional list, each subarray representing an arm-motor combination with 6 parameters given in the following table. All together, a hexacopter is represented by 36 parameters. 

\begin{table}[htbp]

\centering
\begin{tabularx}{\textwidth}{p{1.5cm}p{2cm}X}
\toprule
\textbf{Symbol} & \textbf{Range} & \textbf{Description} \\
\midrule
$l$ & $[0.09, 0.4]$ m & Arm length; motors are placed at the end of the arm, facing upward  along the z-axis.\\
$\psi_a$ & $[0, 2\pi]$ rad & Arm polar rotation angle (anticlockwise).\\
$\theta_a$ & $[-\pi, \pi]$ rad & Arm azimuth rotation angle. Positive z-axis is upward. \\
$\psi_m$ & $[0, 2\pi]$ rad & 
Motor polar rotation angle (anticlockwise). \\
$\theta_m$ & $[-\pi, \pi]$ rad & 
Motor azimuth rotation angle. \\
$dir$ & $\{0, 1\}$ & 
Binary value indicating direction of motor spin.\\
\bottomrule
\end{tabularx}
\label{tab:mutation_params}
\end{table}

Reproduction is conducted by replicating a parent and applying mutation to the offspring\footnote{We do not applied crossover because preliminary experiments with crossover resulted in lower performance. See supplementary material for more details}. Mutation occurs on an individual $ \mathbf{A} $ which is represented as a matrix of \( n \) arms, each with $m$ parameters. First, an arm is selected by uniform selection, and then a parameter is selected for mutation. For nondirectional parameters, sampled Gaussian noise is added, followed by clipping or wrapping to maintain valid bounds (arm lengths are clipped to their limits and angles are wrapped to \([0, 2\pi]\)). For the direction parameter, mutation simply flips its binary value. Mutation is always followed by applying a repair operator to ensure that the mutated structure remains valid and collision-free. Specifically, we prevent rotor overlap by modeling each motor as occupying a three-dimensional bounding box centered on its position. The dimensions of each bounding box are defined by the rotor geometry: the width and depth correspond to the diameter of the propeller \( H \), and the height is set to \( 2H \). 
The increased vertical space accounts for aerodynamic downwash and minimizes rotor interference.

\subsubsection{Initialization and Viability Filtering} \label{sec:init_pop}

The initial population of drone morphologies can be generated randomly. However, this approach often results in a significant number of drones that are incapable of flight due to insufficient thrust to counteract gravity. To avoid wasting computational resources on such non-viable designs, we apply a static hovering test, based on the method described in \cite{ang2025multiobjectiveevolutiondronemorphology}, to filter them out. This ensures that only drones with the minimum capability to hover are included in the initial population. It is important to note that this viability check sets only a baseline threshold and does not imply high performance; the selected morphologies must still evolve to achieve optimal behavior. Preliminary tests showed that only $43\%$ of completely random designs pass the hovering test, and even fewer can learn to accomplish the specific tasks. This highlights the importance of filtering at the population initialization stage.  

\section{Learning process}
\label{sec:rl}

Reinforcement learning is a paradigm in which an agent learns to make decisions by interacting with an environment in order to maximize cumulative rewards. 

We employ Proximal Policy Optimization (PPO) \cite{PPO} that follows an actor-critic framework: a policy network (actor) proposes actions based on the current state, while a value network (critic) estimates the expected return from that state. Training proceeds by collecting \textit{rollouts}, sequences of states, actions, and rewards gathered by running the current policy in the environment. Instead of updating after each individual interaction, PPO accumulates a batch of rollout data and optimizes the policy over several epochs using \textit{mini-batches}, randomly sampled subsets of the collected data. This improves sample efficiency and stabilizes learning. PPO maximizes a clipped surrogate objective that limits the size of policy updates, ensuring more stable and conservative improvements. Its combination of simplicity, robustness, and efficiency makes it particularly suitable for continuous control tasks in robotics.

For all tasks, we use a reward function that reflects the velocity the drone travels towards the next gate, plus some penalty for rotating too much. This penalty is needed to guide the learning process to begin with, so that the drone can learn to hover. The specific formula is $r_t = (s_{t-1} - s_{t}) - 0.001 \cdot \| \boldsymbol{\Omega}_t\|$,

where $r_t$ is the reward at timestep $t$, $s_t$ is the distance to the next gate at timestep $t$ and $\boldsymbol{\Omega}_t$ is the rotational velocity at timestep $t$. The observation the drone receives contains 16 variables: 3 for position of the drone relative to the next gate, 3 for velocity of the drone relative to the next gate, 3 for the Euler angle of the drone, 3 for the angular velocity of the drone, and 4 more for the relative position and yaw of the next gate with respect to the current target gate.

Note that the reward function used for learning and the fitness function to drive evolution are different. This was chosen because fitness is meant to reflect the overall task performance (which is captured well by the number of waypoints), whereas reinforcement learning needs a reward function that provides frequent and informative feedback to guide learning over short timescales.

\section{Metrics for Analysis} \label{sec:metrics}

{\bf Novelty and Diversity Measures in Optimization} Our basic metric is to quantify the difference between two individuals by the \emph{edit distance} in relation to the spatial configuration of their arms. Each individual is represented as a two-dimensional array of parameters, where the first dimension indexes the arms, and the second dimension contains their associated features. Formally, let \( \mathbf{A} = \{ \mathbf{a}_1, \ldots, \mathbf{a}_n \} \) and \( \mathbf{E} = \{ \mathbf{e}_1, \ldots, \mathbf{e}_n \} \) be two such individuals, each with \( n \) arms and \( \mathbf{a}_i, \mathbf{e}_j \in [0, 1]^m \). The edit distance \( D(\mathbf{A}, \mathbf{E}) \) is computed as follows:
\begin{enumerate}
    \item Normalize all arm parameters to be within \([0, 1]\).
    \item Compute the pairwise cost matrix \( \mathbf{C} \in \mathbb{R}^{n \times n} \) where $C_{ij} = \| \mathbf{a}_i - \mathbf{e}_j \|_2$ is the Euclidean distance between arm $\mathbf{a}_i$ from $\mathbf{A}$ and arm $\mathbf{e}_j$ from $\mathbf{E}$.
    \item Solve the assignment problem using the Hungarian algorithm \cite{HungarianAlgorithm} to find a bijection $ \pi: \{1, \ldots, n\} \rightarrow \{1, \ldots, n\} $ that minimizes total cost:
    \begin{equation}
    D(\mathbf{A}, \mathbf{E}) = \sum_{i=1}^n \| \mathbf{a}_i - \mathbf{e}_{\pi(i)} \|_2
    \end{equation}
\end{enumerate}

The \emph{novelty} of an individual \( \mathbf{A} \) w.r.t. a set of individuals \( \mathbf{S} = \{\mathbf{E}^{(1)}, \ldots, \mathbf{E}^{(\mu)} \)\} is calculated as the normalized average of the pairwise edit distances:
\begin{equation}
\text{N}(\mathbf{A},\mathbf{S}) = \frac{1}{\mu} \sum_{j=1}^\mu \frac{D(\mathbf{A}, \mathbf{E}^{(j)})}{D_{\max}}
\label{eq:novelty}
\end{equation}
where \( D_{\max} \) is the maximum pairwise edit distance observed in the reference set \( \mathbf{S} \), used for normalization.

The \emph{diversity} of a set of individuals \( \mathbf{S} = \{\mathbf{E}^{(1)}, \ldots, \mathbf{E}^{(\mu)} \)\} is defined as the average edit distance between the members of that set:
\begin{equation}
\text{Div}(\mathbf{S}) = \frac{1}{\mu} \sum_{i=1}^\mu \sum_{j=1}^\mu \text{D}(\mathbf{E}^{(i)},\mathbf{E}^{(j)})
\end{equation}

\textbf{Learning Dynamics Evaluation Metrics} To understand how morphology affects learning, and how learning, in turn, shapes morphology through evolution, we require metrics that capture distinct aspects of the learning process. 

\begin{figure}[hbt!]
    \centering
    \includegraphics[width=0.85\linewidth]{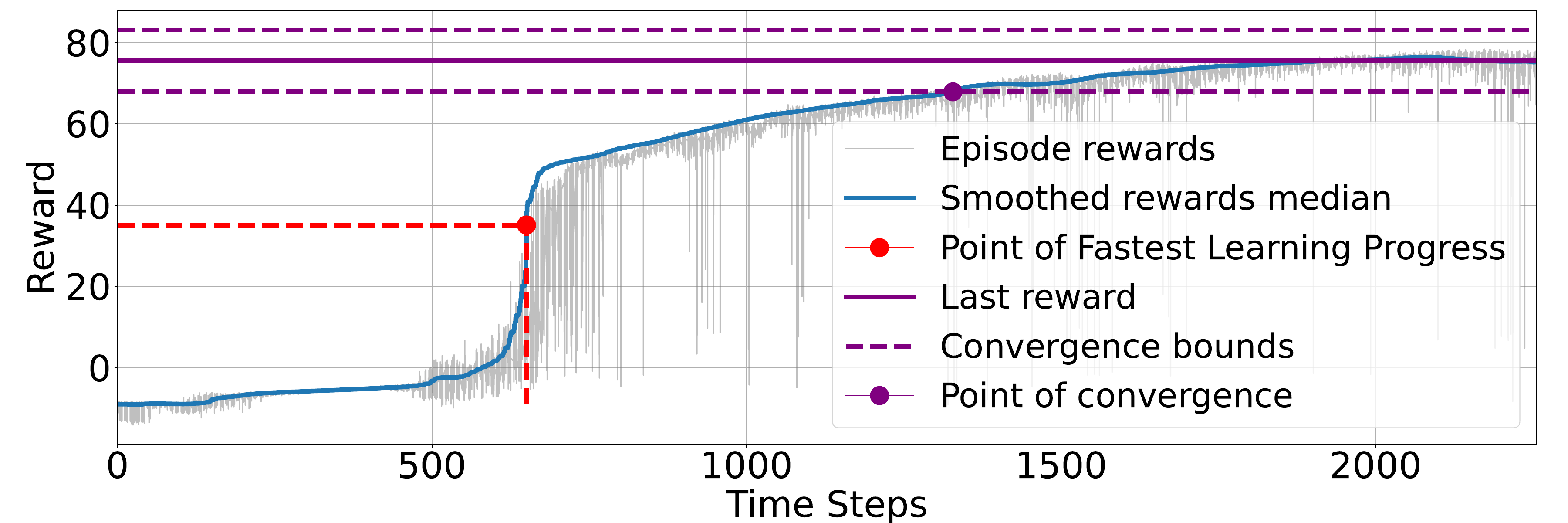}
    \caption{Example learning curve 
    }
    \label{fig:learing_descriptors_example}
\end{figure}

To this end, we define a series of quantitative metrics based on the episodic reward signal. These metrics are computed based on a smoothed version of episodic rewards, denoted $\tilde{r}_t$, obtained using a median sliding window aggregation over a fixed-size window of $1000$ episodes. The key idea behind these metrics is to define two important moments during the learning process and divide the full learning period into three stages accordingly: the early stage, the mid stage, and the late stage. Figure~\ref{fig:learing_descriptors_example} illustrates the main concepts visually.

The first stage represents the period before the most rapid progress in learning, when the rewards are most rapidly increasing. We define this point as the \textbf{Point of Fastest Learning Progress} ($t_b$), which is the time step at which the maximum positive change in the smoothed reward curve occurs: $ t_b = \underset{t}{\arg\max} \, \Delta \tilde{r}_t$, where $\Delta \tilde{r}_t = \tilde{r}_{t+1} - \tilde{r}_t$ denotes the first difference of the smoothed reward series.

This point marks a critical transition in the learning process. Since our reward function reflects the velocity the drone travels towards the next gate, this point typically corresponds to when the agent discovers fundamental flight behaviors, transitioning from random exploration to acquiring basic competency in the task. The timing of this breakthrough is important for understanding the morphological effects on learning: morphologies that enable earlier discovery of effective behaviors demonstrate better compatibility with the learning process.

The second important moment is the \textbf{Point of Convergence} ($t_c$), defined as the earliest timestep after which the smoothed rewards remain within a fixed relative tolerance ($\pm 10\%$) of the final reward $\tilde{r}_T$. Formally, this is not the end of learning, but improvements after this point are relatively small. One could say that after this point, drones are only fine-tuning their behavior. 

Next, we define a metric to reflect the progress made between the point of fastest learning progress and the point of convergence. The \textbf{Learning Speed} ($s$) is quantified as the maximum reward achieved during the mid stage of the learning period divided by the duration of that period: $s = \frac{\tilde{r}_{\mathrm{max}}}{t_c - t_b}$.

Note that learning speed is a relative metric because the maximum reward is pertained to the length of a time interval. Therefore, it reflects the efficiency of learning. An absolute measure to evaluate learning efficacy is the \textbf{Maximum Reward Achieved}, defined as the maximum smoothed episodic reward achieved during training: $\tilde{r}_{\mathrm{max}} = \max_{t} \tilde{r}_t$.

The last metric we introduce is not related to efficiency, nor to efficacy. Instead, it indicates how stable the learning process is. We measure \textbf{Learning Stability} ($\sigma_{\Delta \tilde{r}}$) by the mean absolute difference between consecutive smoothed episodic rewards:
\begin{equation}
    \sigma_{\Delta \tilde{r}} = \frac{1}{T-1} \sum_{t=0}^{T-2} \left| \tilde{r}_{t+1} - \tilde{r}_t \right|.
\end{equation}

\section{Experiments}

Experiments were conducted in a custom multirotor simulator implemented in Python 3. The simulator follows the principles outlined by Ferede et al. \cite{ferede2025netrulealldomain}.\footnote{Details are provided in the supplementary material. \label{fn:sup_material}}

The following properties were held constant: the mass and inertia of the central unit, the radius and thrust coefficient of individual propellers, and the density of structural arms\footref{fn:sup_material}.
Starting positions were fixed across trials to reduce computational complexity. Although domain randomization can promote robustness, it was not employed here to maintain manageable training times.

Each evolutionary run used $\mu = 24$ and $\lambda = 24$ over 40 generations, yielding 960 individuals in total. Each new morphology created by reproduction performed a learning process through proximal policy optimization (PPO) using the implementation of Stable-Baselines3~\cite{SB3}. The computational budget for learning a policy was $2.5 \times 10^8$ time steps. This value was chosen because it was the minimum number of training steps needed for the conventional hexacopter design to learn the tasks.\footref{fn:sup_material} Training and evaluation of a single controller took approximately ten minutes. To ensure robust results, each experiment was replicated 10 times. 

Post experimental analysis targeted three aspects of the optimization process, the evolutionary dynamics, the quality of the evolved designs, and the learning dynamics. To supplement the quantitative data, video demonstrations of both the baseline hexacopter and evolved drones are available on YouTube\footnote{https://youtu.be/playlist?list=PL5oQiyJFx4qM9Hzs2asyoGbJo9TuO4sPS} providing visual documentation of the emerged flight behaviors.

\subsubsection{Evolutionary Dynamics} \label{sec:progression_of_the_evolutionary_optimization}

\begin{figure}[th!]
\centering
 \subcaptionbox{Circle}{
	\includegraphics[width=0.22\linewidth]{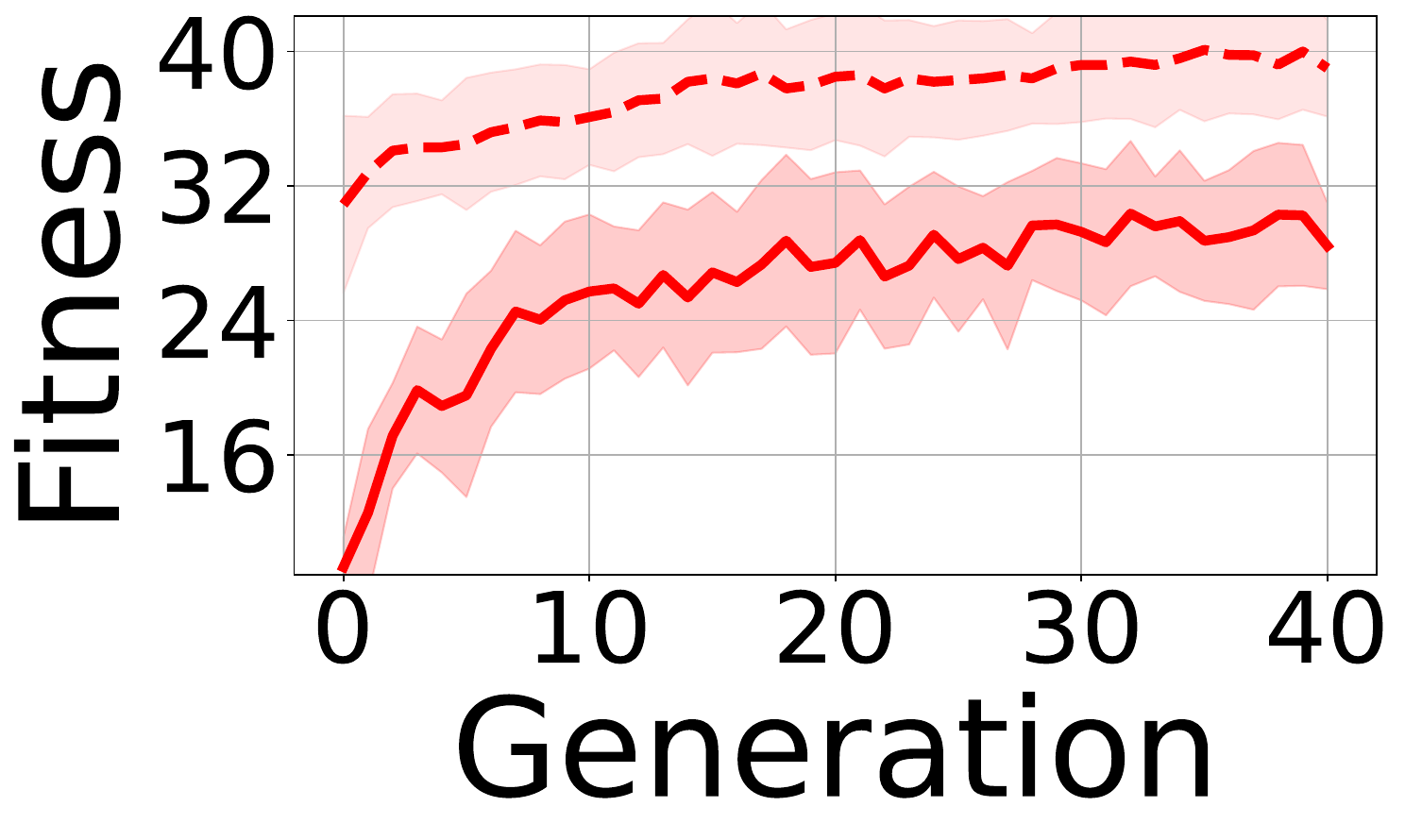}
 }
 \subcaptionbox{Slalom}{
	\includegraphics[width=0.22\linewidth]{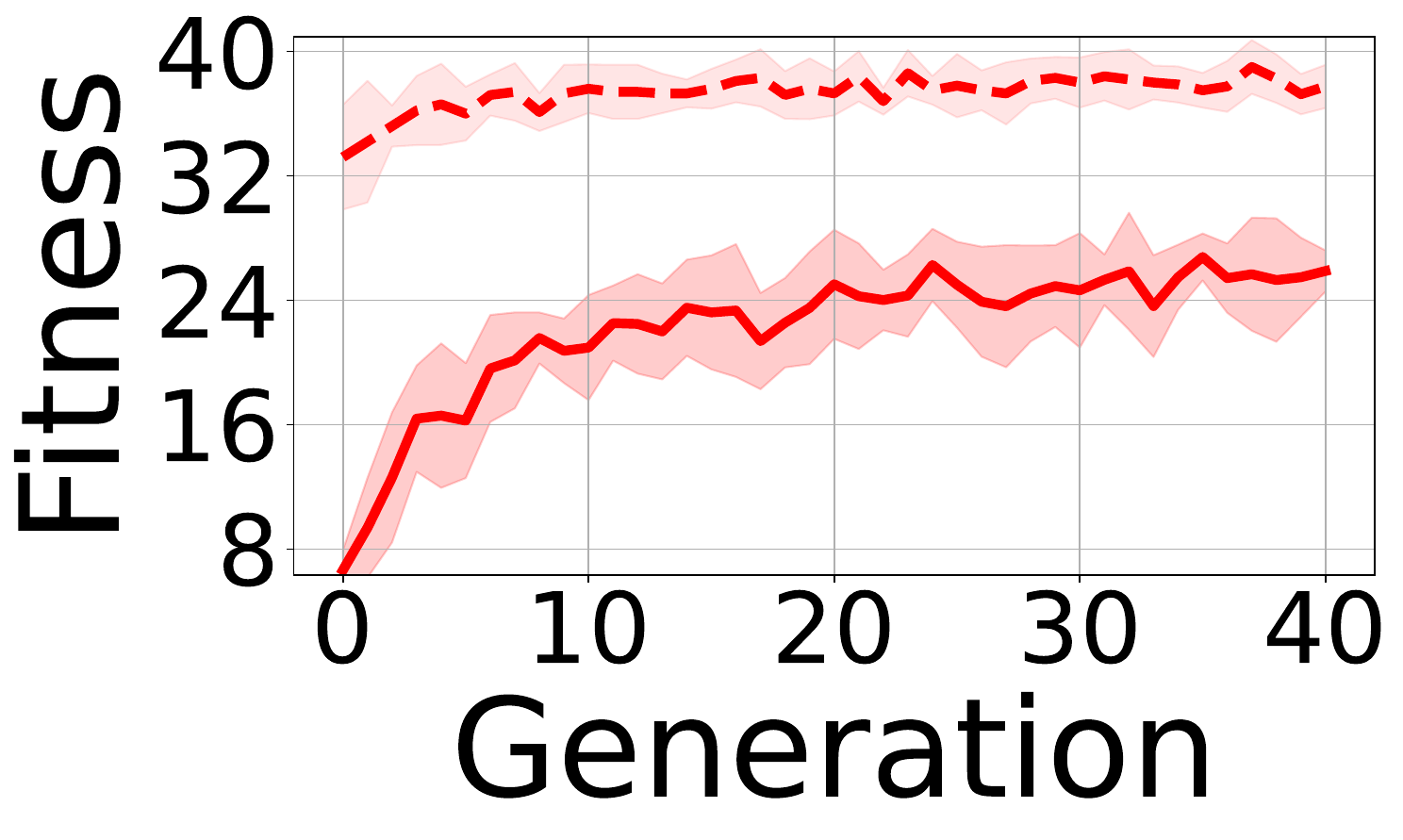}
 }
 \subcaptionbox{Shuttleruns}{
	\includegraphics[width=0.22\linewidth]{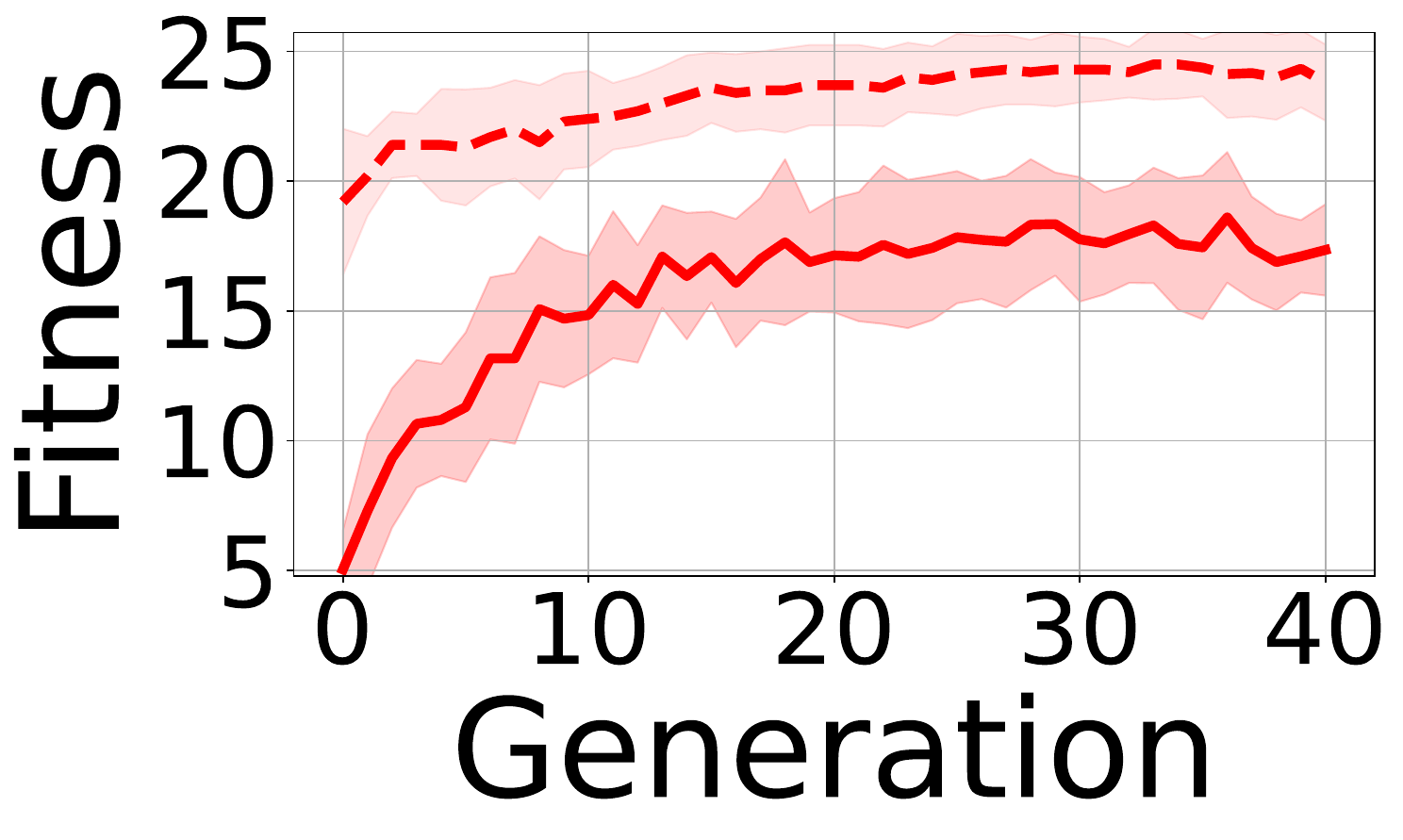}
 }
 \subcaptionbox{Figure8}{
	\includegraphics[width=0.22\linewidth]{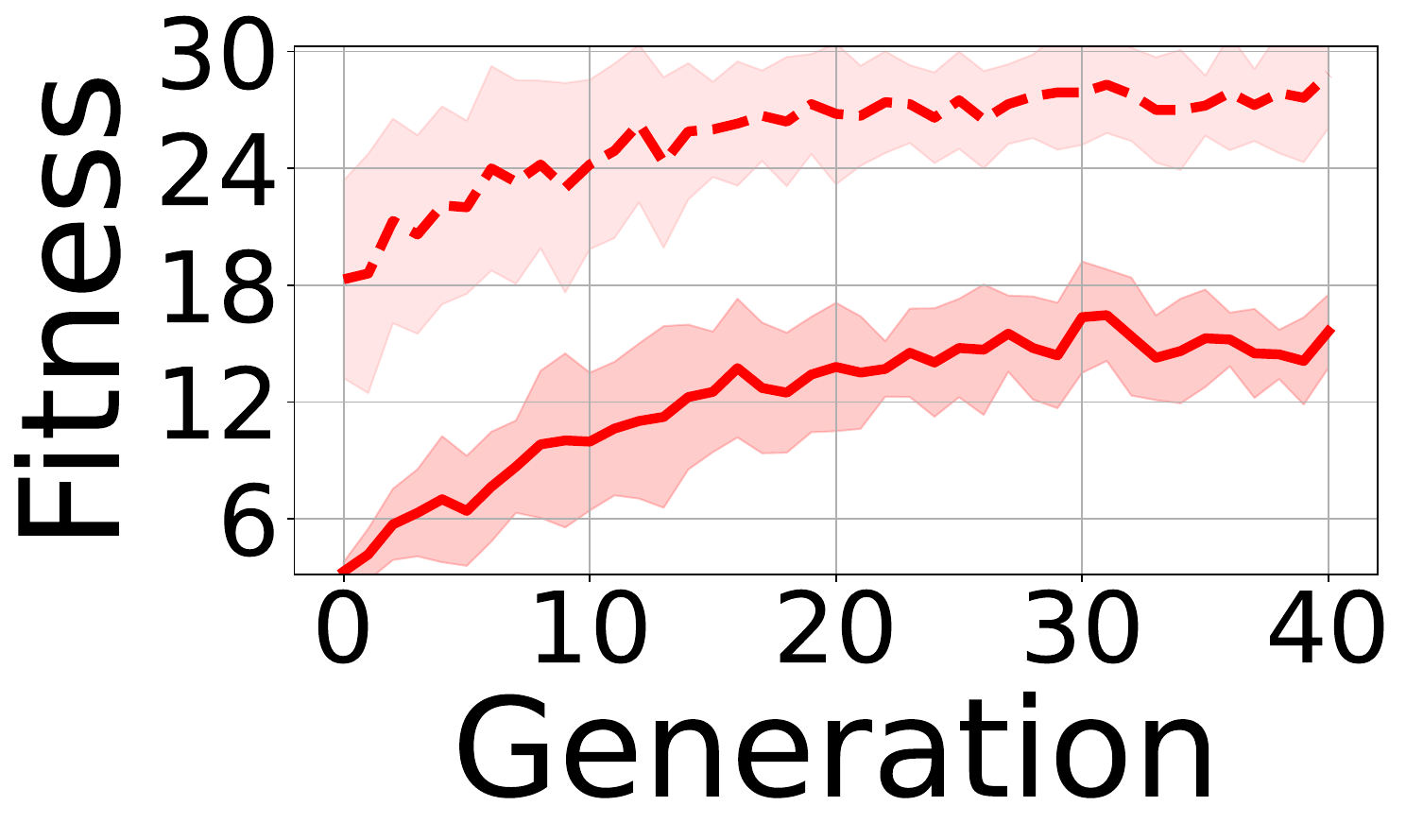}
 }
  \subcaptionbox{Circle}{
	\includegraphics[width=0.22\linewidth]{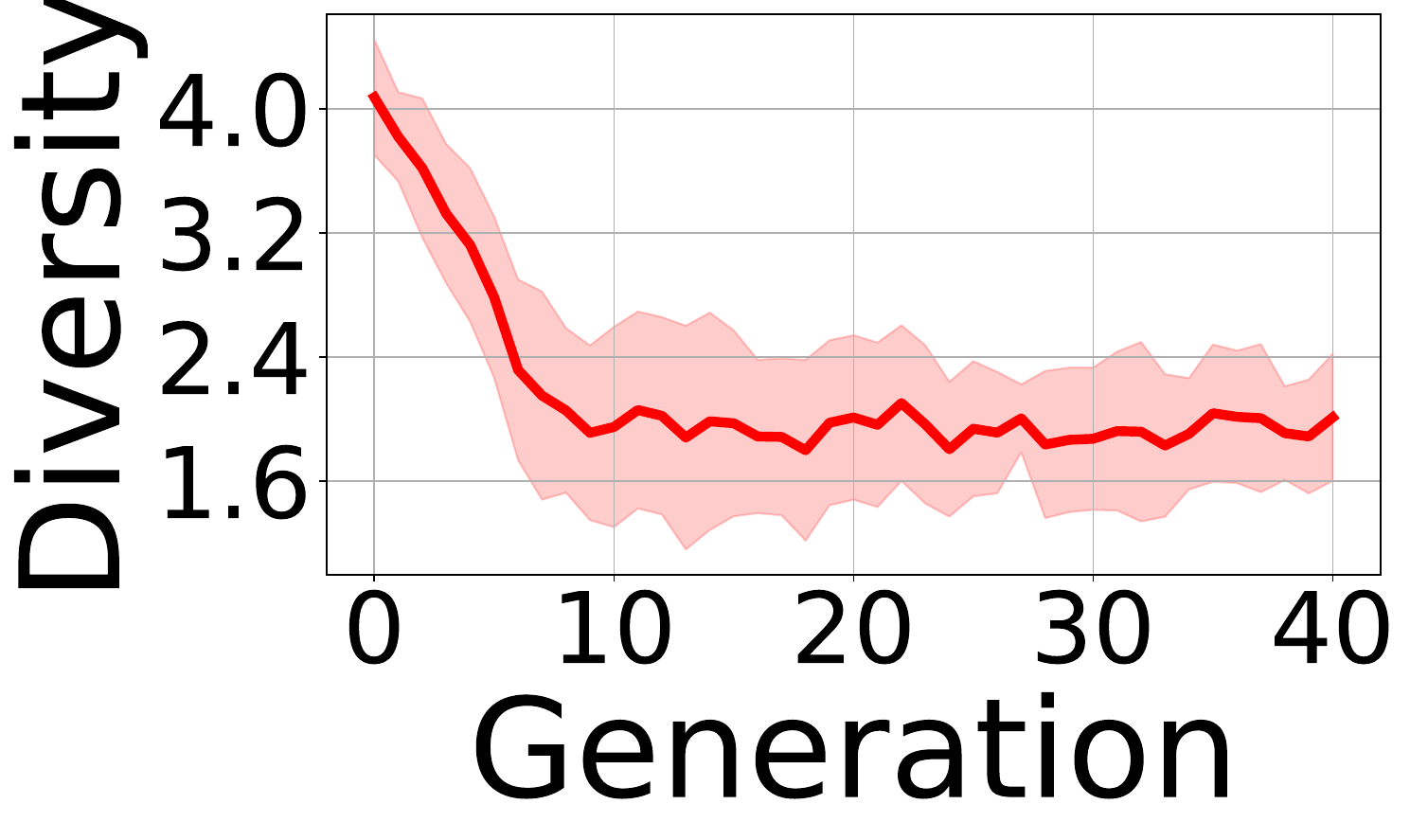}
 }
 \subcaptionbox{Slalom}{
	\includegraphics[width=0.22\linewidth]{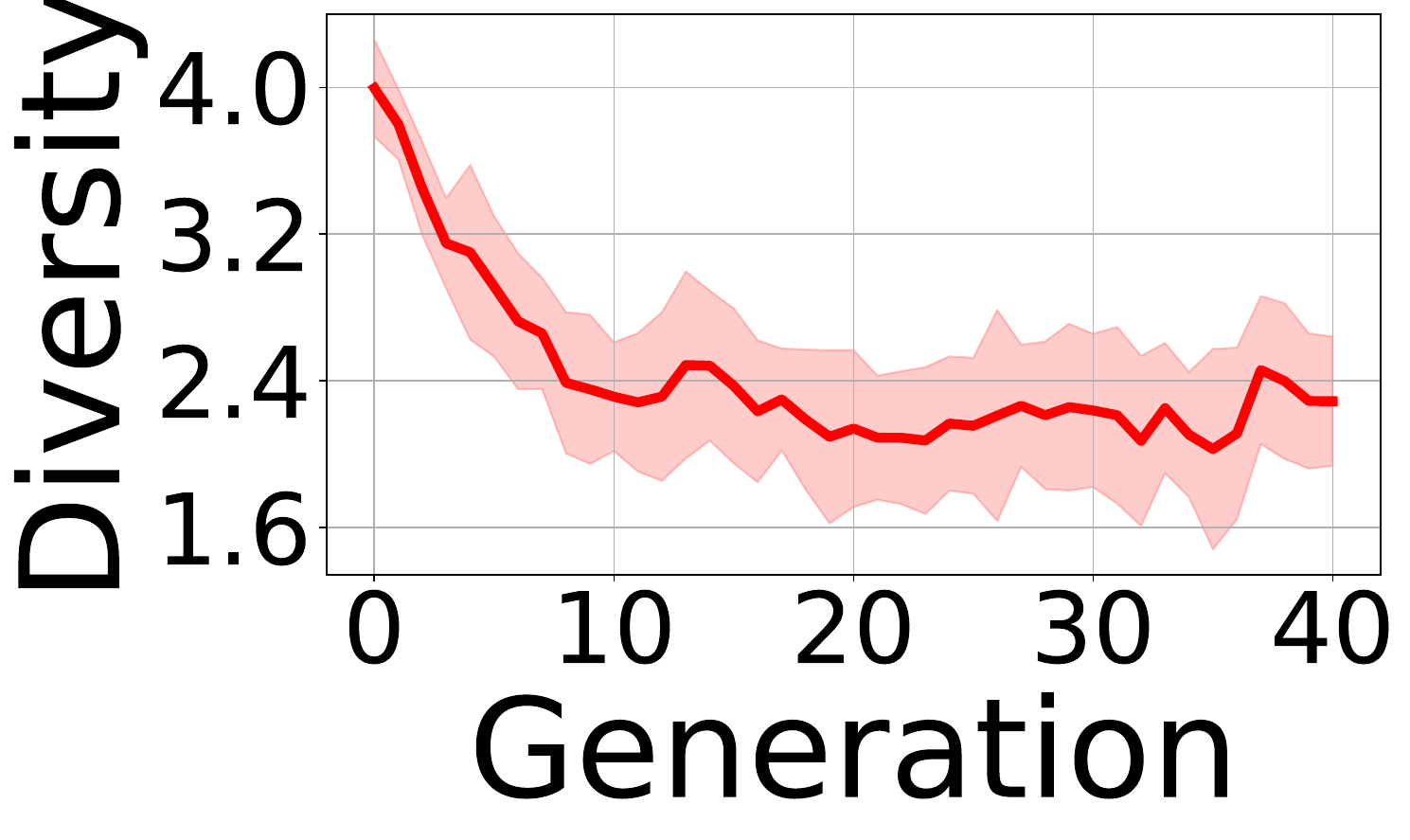}
 }
 \subcaptionbox{Shuttleruns}{
	\includegraphics[width=0.22\linewidth]{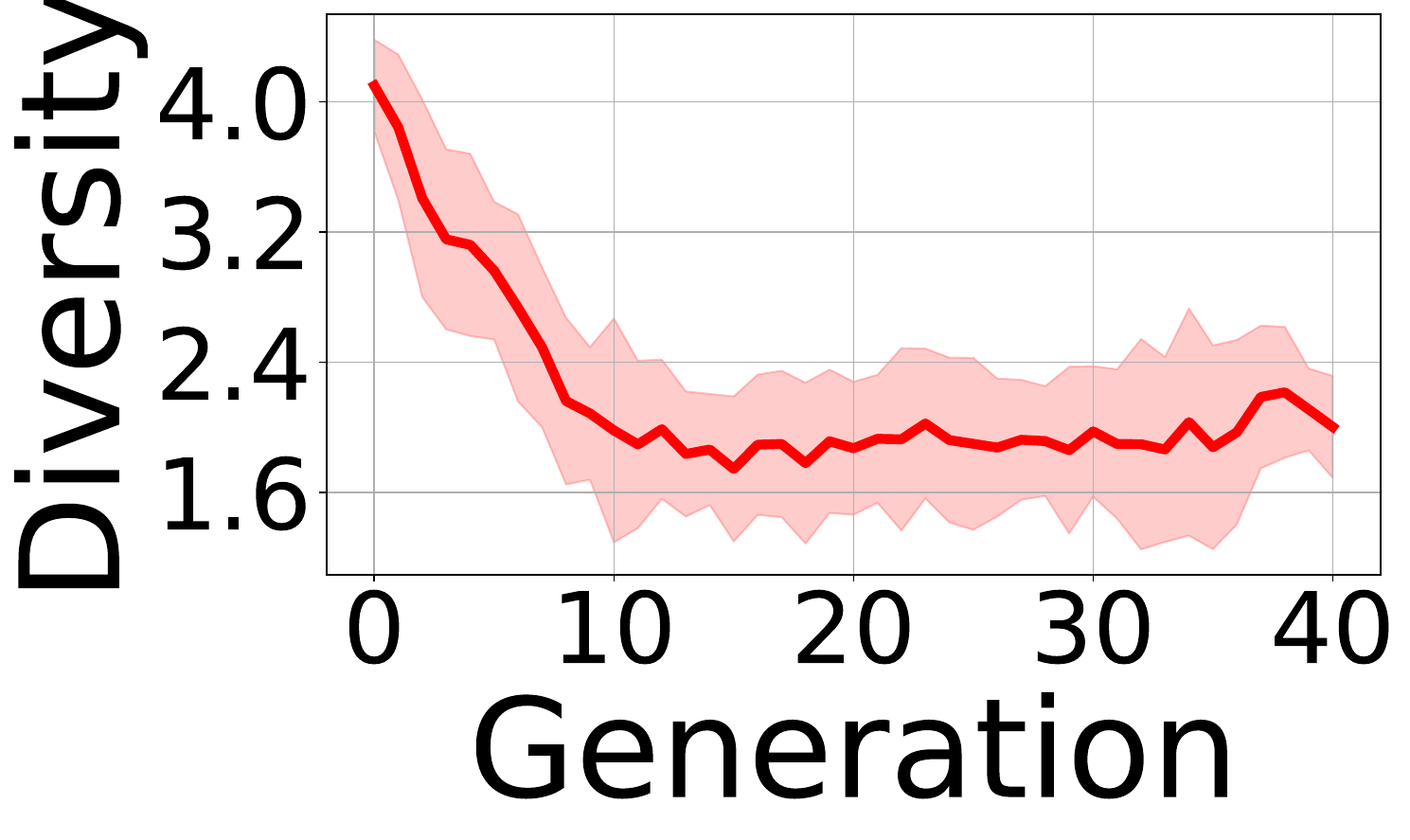}
 }
 \subcaptionbox{Figure8}{
	\includegraphics[width=0.22\linewidth]{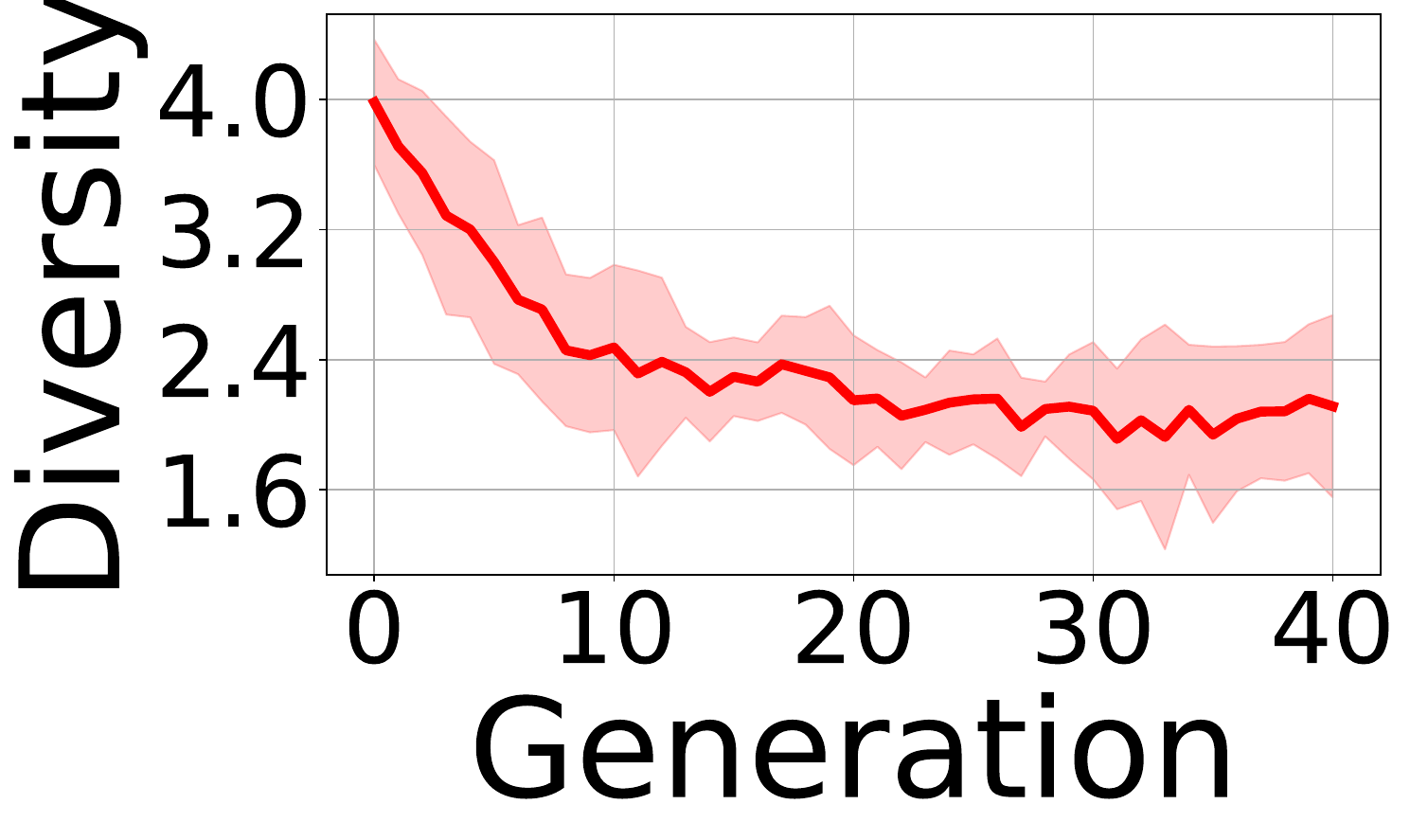}
 }
\caption{Fitness and diversity results averaged over 10 runs. The top row shows the fitness plots, where solid lines represent the average fitness, dotted lines indicate the maximum fitness. The bottom row displays the change of diversity over generations. For all curves, the shaded areas show the standard deviation.}
\label{fitness_diversity_plots}
\end{figure}

Figure \ref{fitness_diversity_plots} illustrates the progression of fitness and population diversity across generations. For all tasks, we observe a steady improvement in average fitness up to approximately generation 20, after which the curves plateau with only marginal changes. The slalom and circle tasks show relatively modest gains, whereas the shuttle run and figure-eight tasks exhibit clear improvements in both average and maximum fitness. This suggests that a more intricate body design may only provide advantages in more demanding tasks, or that the evolutionary system requires task specific fine tuning. 

The diversity plots reflect the typical dynamics of evolutionary processes: population diversity declines in all tasks, yet a certain degree of variation is retained by the end. Preserving diversity is generally desirable, as it leaves room for further evolutionary progress. Nevertheless, the vast search space and the substantial computational cost of evaluations limit the number of designs that can be tested, which in turn makes the final outcomes strongly dependent on the initial population.

We emphasize that our method is not claimed to be optimal. However, the results clearly demonstrate that the algorithm can discover improved designs across all four tasks.

\subsubsection{Evaluation of Generated Designs}

Inspecting the generated designs we can address the first research question: \textit{how do the evolved drones compare to the usual design?} We break this down into two related subquestions: \textit{how do they perform and how do they look?}
\begin{figure}[hbt!]
\centering
\begin{minipage}[b]{0.45\textwidth}
    \centering
    \includegraphics[width=\linewidth]{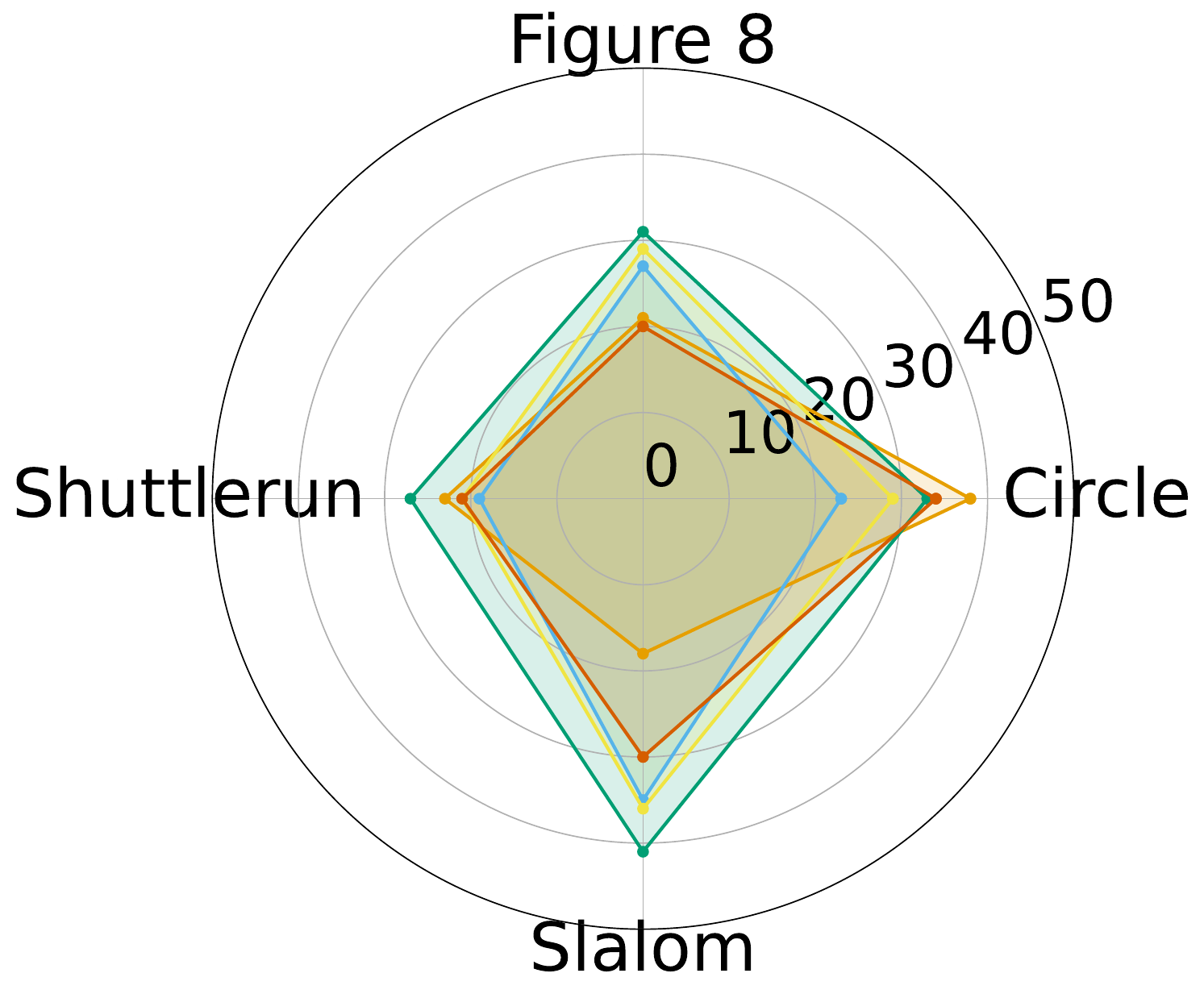}
    \refstepcounter{subfigure}
    \textbf{(a)} Maximum fitness radar plot 
    \label{fig:max_fitness_radar_plot}
\end{minipage}
\begin{minipage}[b]{0.45\textwidth}
    \centering
    \includegraphics[width=\linewidth]{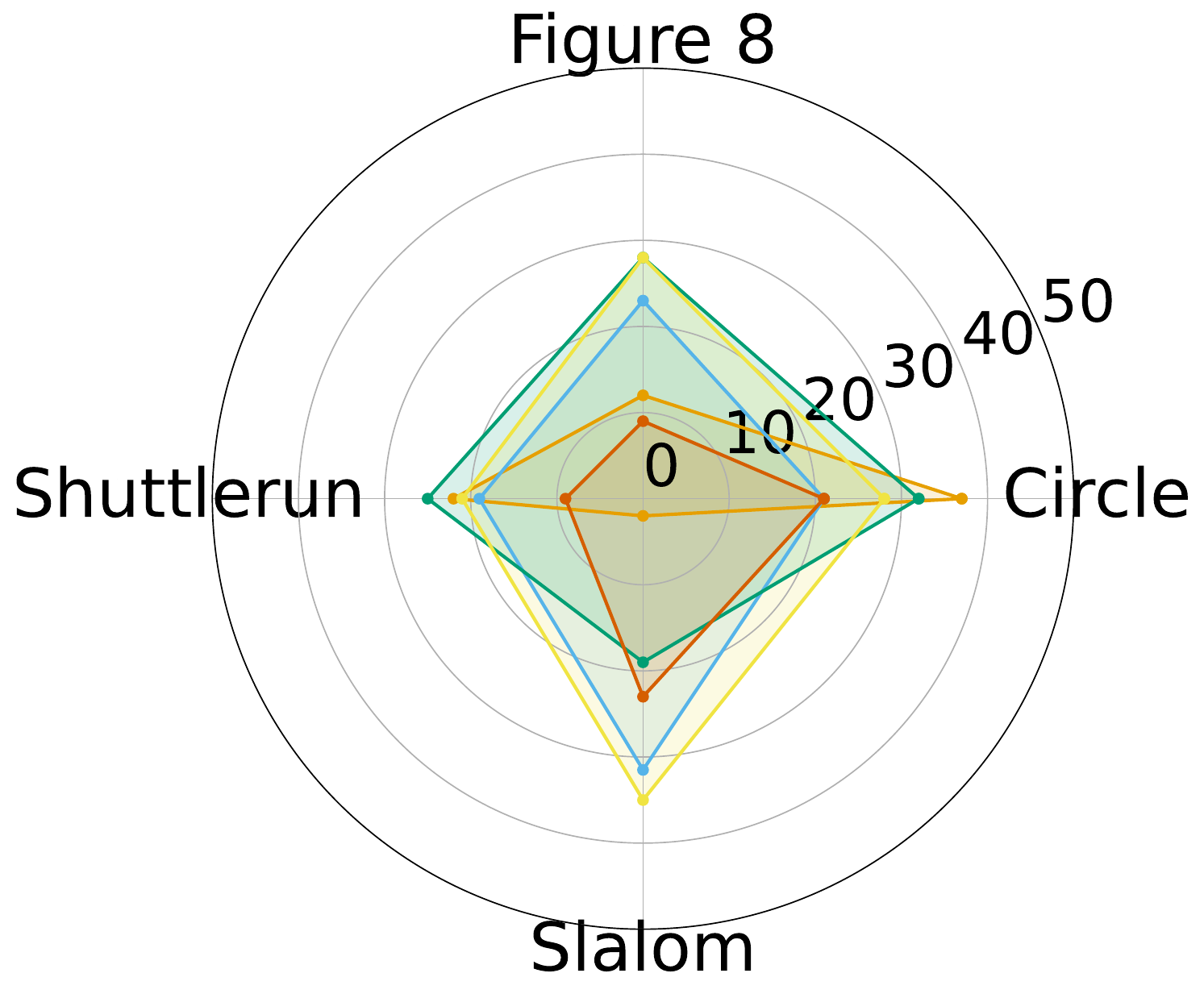}
    \refstepcounter{subfigure}
    \textbf{(b)} Average median fitness radar plot
    \label{fig:avr_fitness_radar_plot}
\end{minipage} 

\begin{subfigure}[b]{0.8\textwidth}
    \centering
    \includegraphics[width=\linewidth]{images/boxplot_legend.pdf}
    \label{fig:radar_legend}
\end{subfigure}
\caption{Maximum and median fitness values achieved by each design per task.}
\label{fig:fitness_radar_plot}
\end{figure}

\begin{figure}[hbt!]
\centering

\begin{subfigure}[b]{0.68\textwidth}
    \centering
    \includegraphics[width=\linewidth]{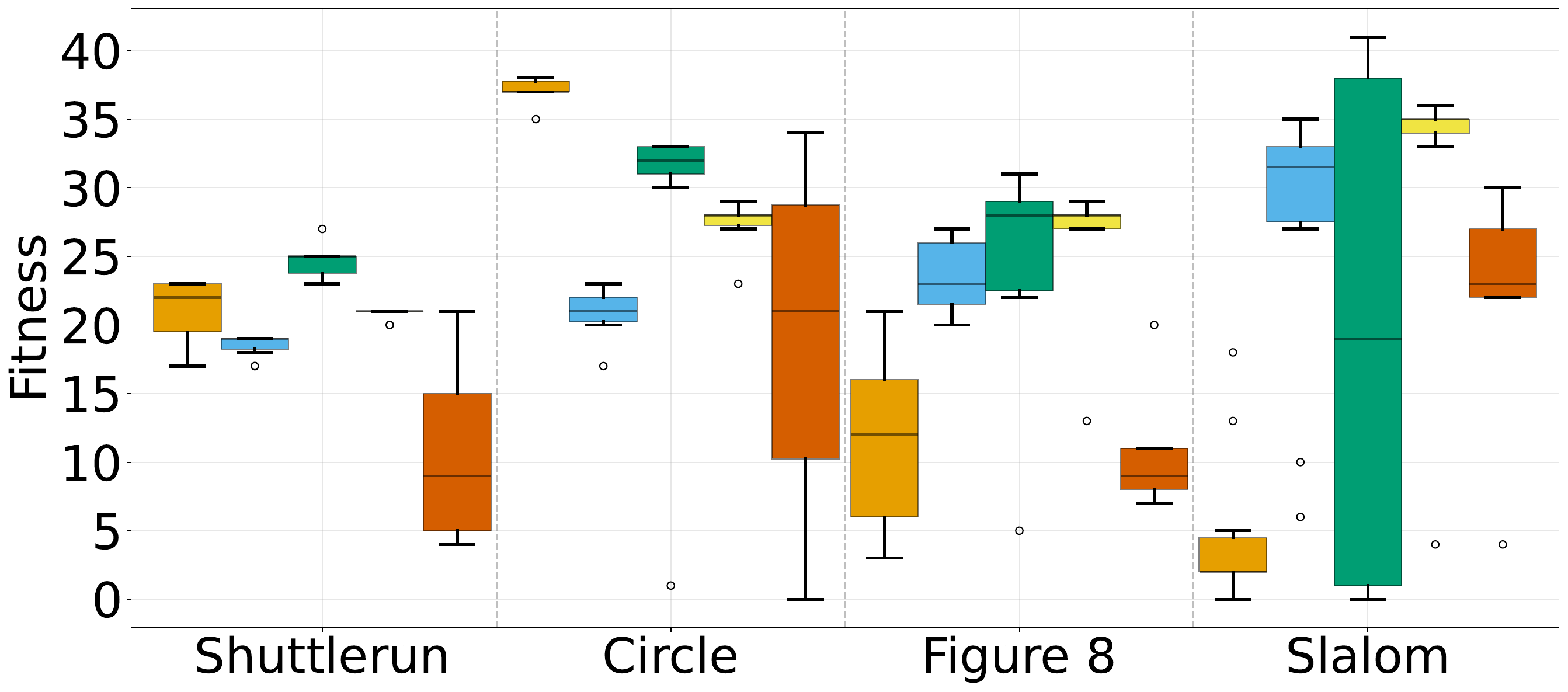}
    \caption{Fitness (high is good)}
    \label{fig:fitness_boxplot}
\end{subfigure}
\begin{subfigure}[b]{0.68\textwidth}
    \centering
    \includegraphics[width=\linewidth]{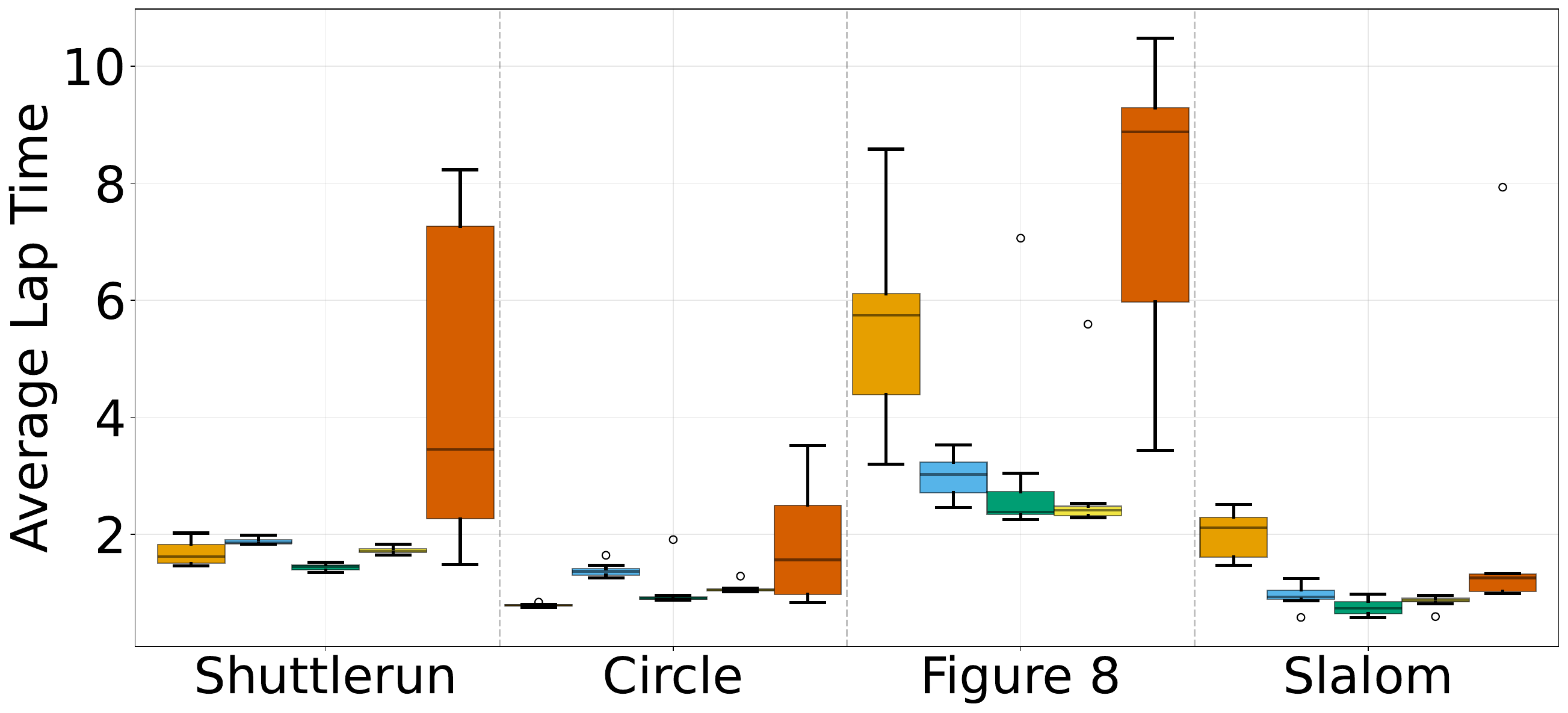}
    \caption{Average lap time (low is good)}
    \label{fig:avg_lap_time_boxplot}
\end{subfigure}
\begin{subfigure}[b]{0.68\textwidth}
    \centering
    \includegraphics[width=\linewidth]{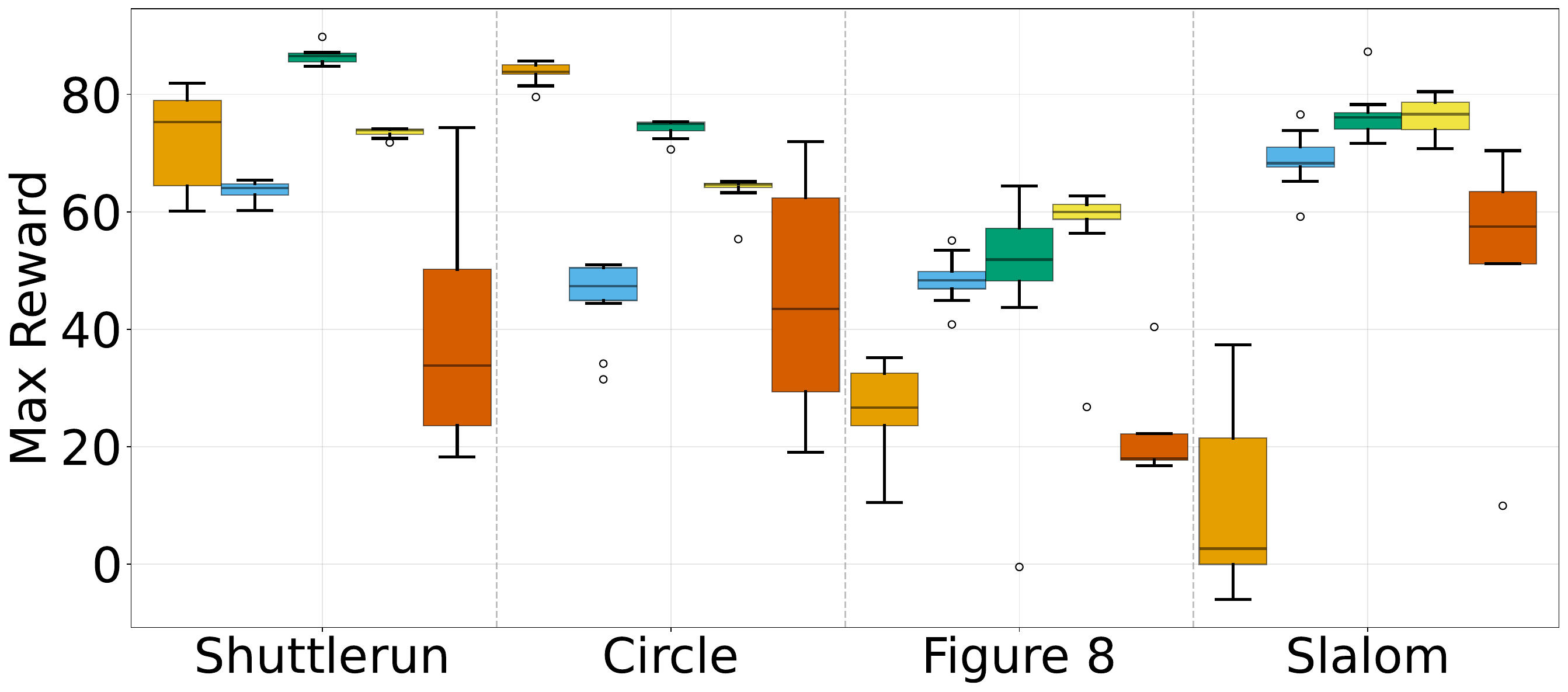}
    \caption{Max reward (high is good)}
    \label{fig:max_reward_boxplot}
\end{subfigure}
\begin{subfigure}[b]{0.68\textwidth} 
    \centering
    \includegraphics[width=\linewidth]{images/boxplot_legend.pdf}
    \label{fig:boxplot_legend}
\end{subfigure}
\vspace{-20pt}
\caption{Performance comparison between traditional and evolved task specialized hexacopters via boxplots showing fitness (gates passed), average lap time and maximum reward achieved. Each morphology was trained and evaluated 10 times.}
\label{fig:box_plots}
\vspace{-15pt}
\end{figure}

Figures \ref{fig:fitness_radar_plot} and \ref{fig:box_plots} summarize the preformance results. Overall, the evolved designs tend to outperform the traditional hexacopter not only on the tasks they were optimized for but also on each other tasks. Interestingly, a design evolved for one specific task often performs better than the standard hexacopter even on tasks it was not optimized for. This general pattern is not without exceptions. For instance, the circle optimized design performs worse than the standard hexacopter on the slalom and figure-eight tasks, while the figure-eight design does not outperform on the circle task. Similarly, on the circle task, the standard hexacopter can perform almost as well as the slalom design. These outcomes can be explained by the behavioral specializations of each morphology. The circle design, tuned for continuous turning in a single direction, struggles on tasks that require alternating turns, such as the slalom and figure eight. Conversely, the figure-eight and slalom designs are optimized for balanced bidirectional turning, which limits their efficiency on a purely unidirectional task like the circle. 

An interesting insight from the results is that generalizability is not compromised by task specific evolution. In fact, the shuttlerun design shows the highest performance across 3 out of 4 tasks, demonstrating remarkable adaptability. We attribute this to the skill set needed for the shuttlerun: rapid acceleration, rapid deceleration, and turning. These skills are broadly useful across racetrack style tasks, making the design highly transferable. Another notable finding is that the shuttlerun design outperforms the figure-eight and slalom designs on their own tasks. However, while the shuttlerun design can achieve strong performance on these tasks, the variance in learning success is high. This means that although the best policies derived from the shuttlerun morphology outperform the figure-eight and slalom designs, achieving such performance is difficult. Since our optimization process favors designs that are consistently learnable, it is naturally biased away from less learnable regions of the design space. 

Figure~\ref{fig:fitness_radar_plot} shows the maximum fitness and the median fitness of each of the designs. The discrepancy between the two plots is an indication of the uncertainty of the fitness of each design. For example,  figure~\ref{fig:fitness_radar_plot}a, shows the generally high performance of the Shuttlerun design. However, figure~\ref{fig:fitness_radar_plot}b reveals that its performance for the Slalom task is highly volatile, scoring a lower median than that for the standard hexacopter. On the other hand, the Slalom design shows remarkable robustness across all four tasks, showing little difference between the two radar plots.

\vspace{-10pt}
\begin{figure}[hbt]
\centering
\begin{minipage}[b]{0.42\textwidth}
    \centering
    \includegraphics[width=0.48\linewidth]{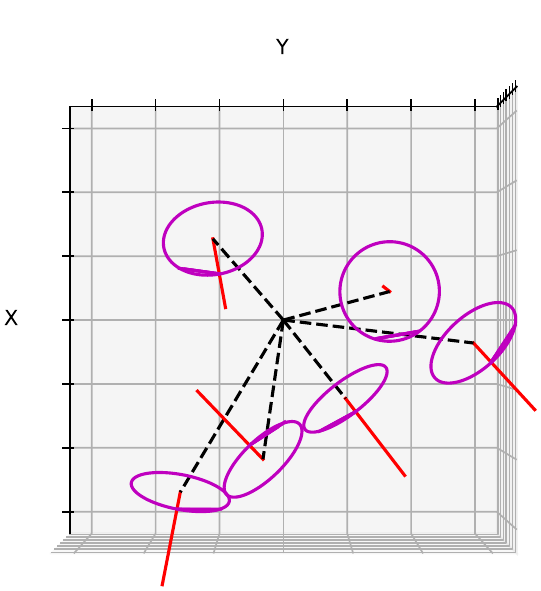}
    \hfill
    \includegraphics[width=0.48\linewidth]{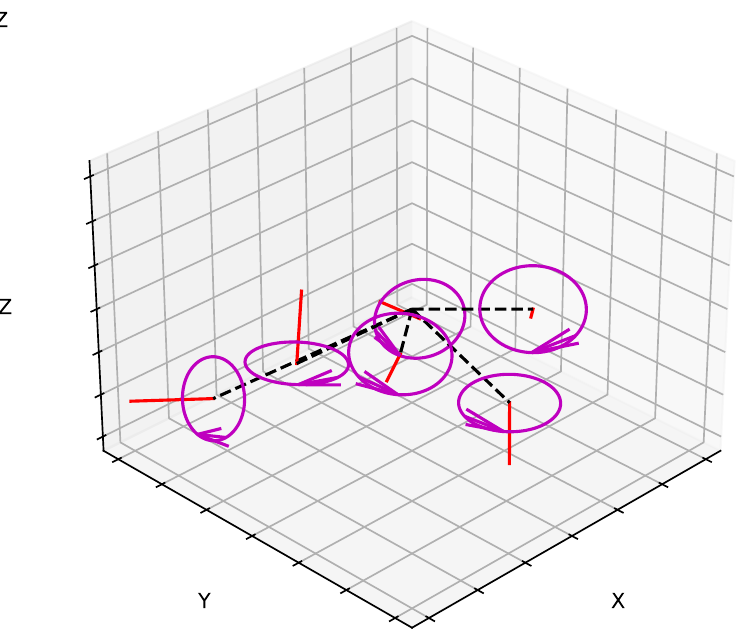}
    \caption{The best evolved drone for the circle task, top view (left) and isometric view (right).}\label{best_inds}
\end{minipage}
\hspace{1em}
\begin{minipage}[b]{0.53\textwidth}
    Figure \ref{best_inds} is an example of the irregular drone designs acquired for each task. Despite differences between tasks, all high performing designs exhibit a pronounced irregularity. The morphologies are asymmetric and unconventional, and no clear intuitive structure is apparent. Notably, all other high performing designs for each task display the same qualitative irregularity, despite differing in their specific layouts.
\end{minipage}
\vspace{-30pt}
\end{figure}

\begin{figure}[h]
\centering
 \subcaptionbox{Circle}{
	\includegraphics[width=0.35\linewidth]{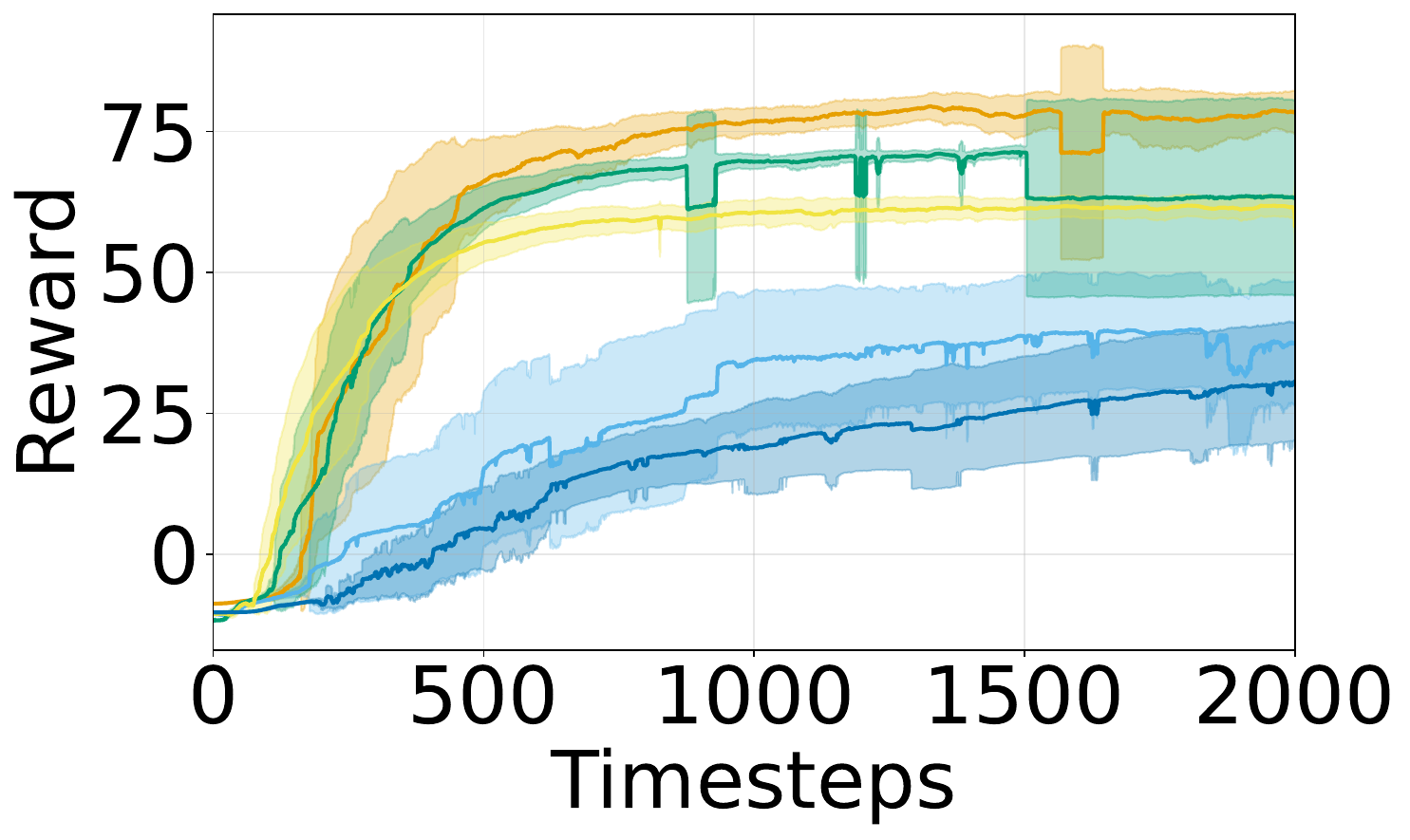}
 }
 \subcaptionbox{Slalom}{
	\includegraphics[width=0.35\linewidth]{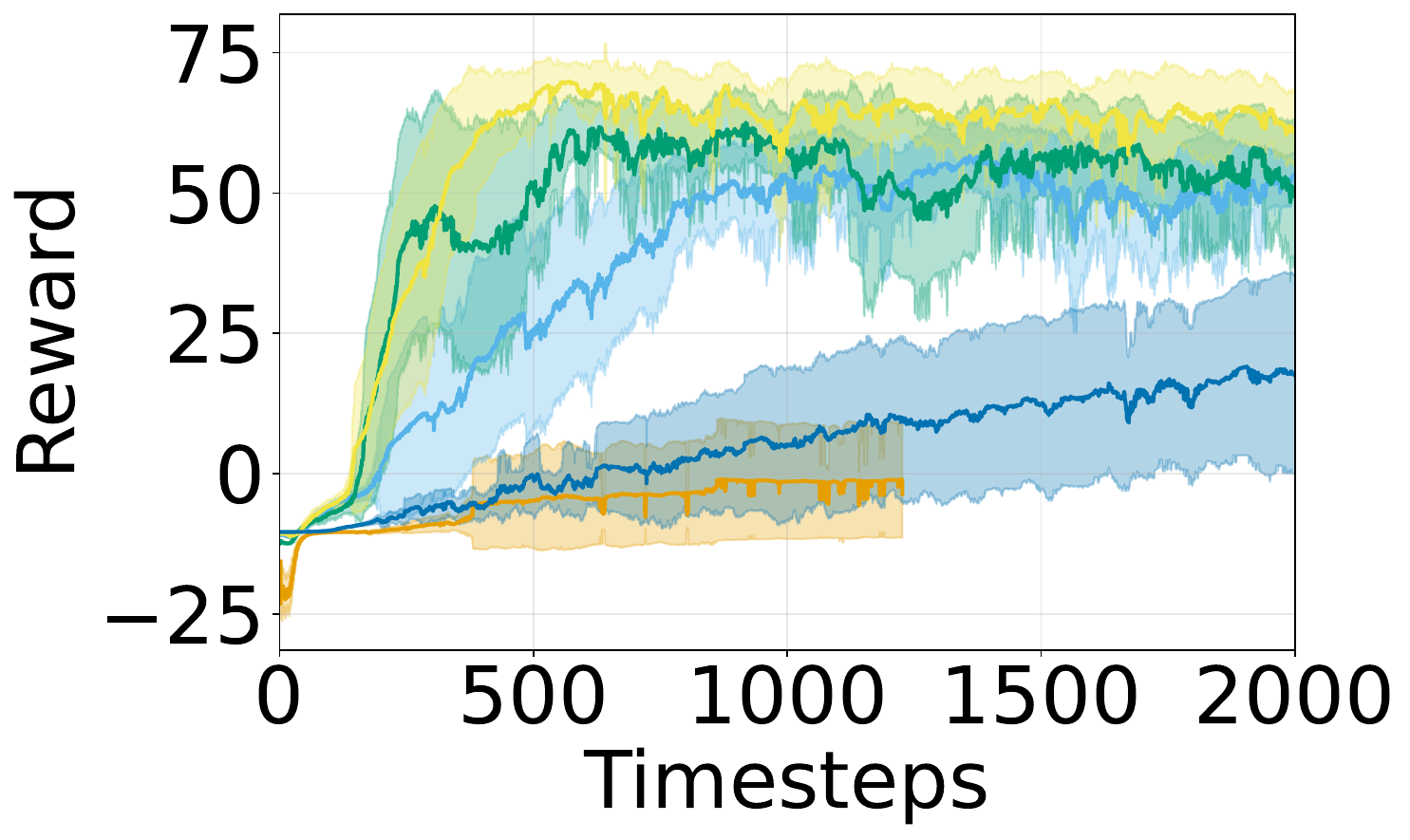}
 }
 \subcaptionbox{Shuttleruns}{
	\includegraphics[width=0.35\linewidth]{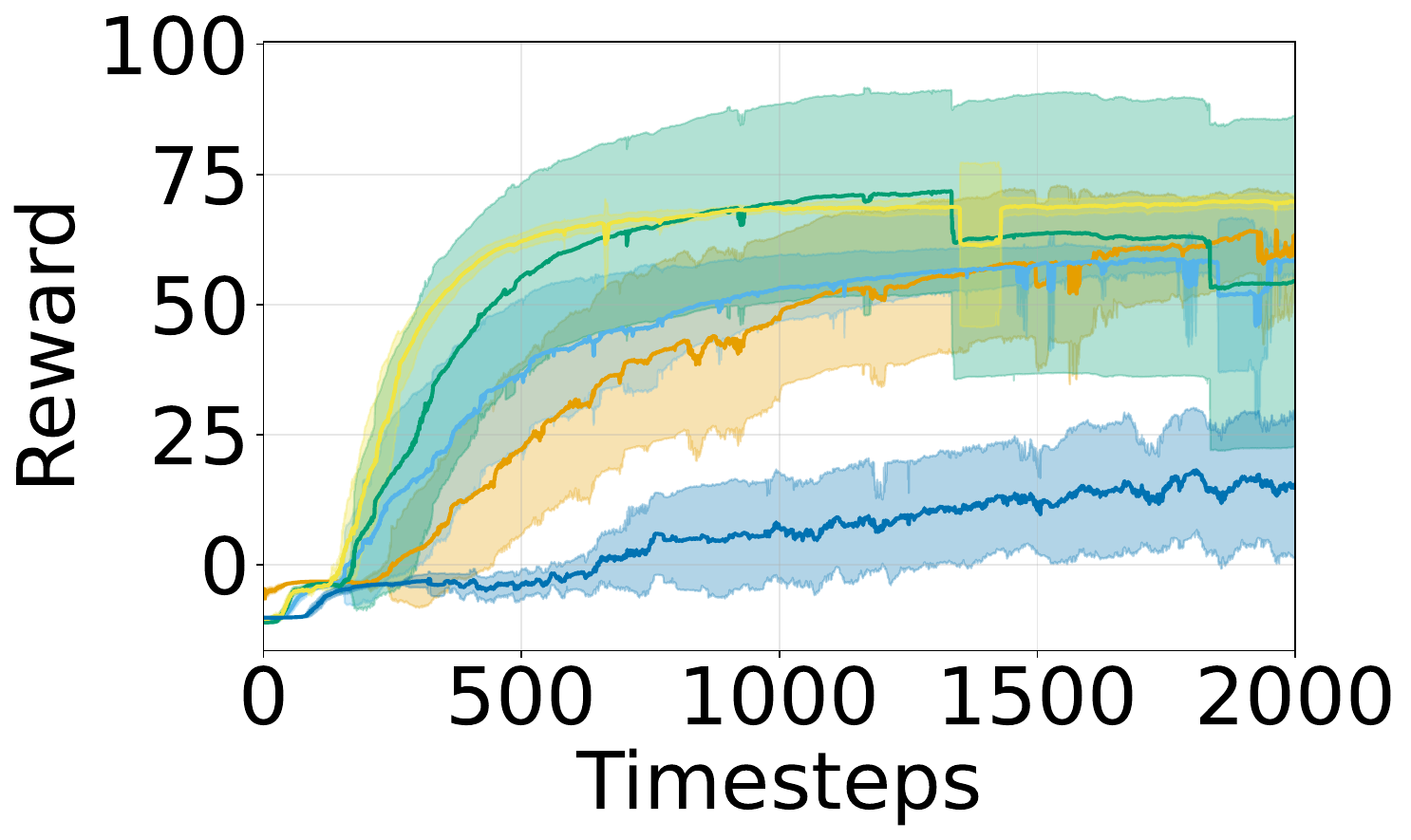}
 }
 \subcaptionbox{Figure8}{
	\includegraphics[width=0.35\linewidth]{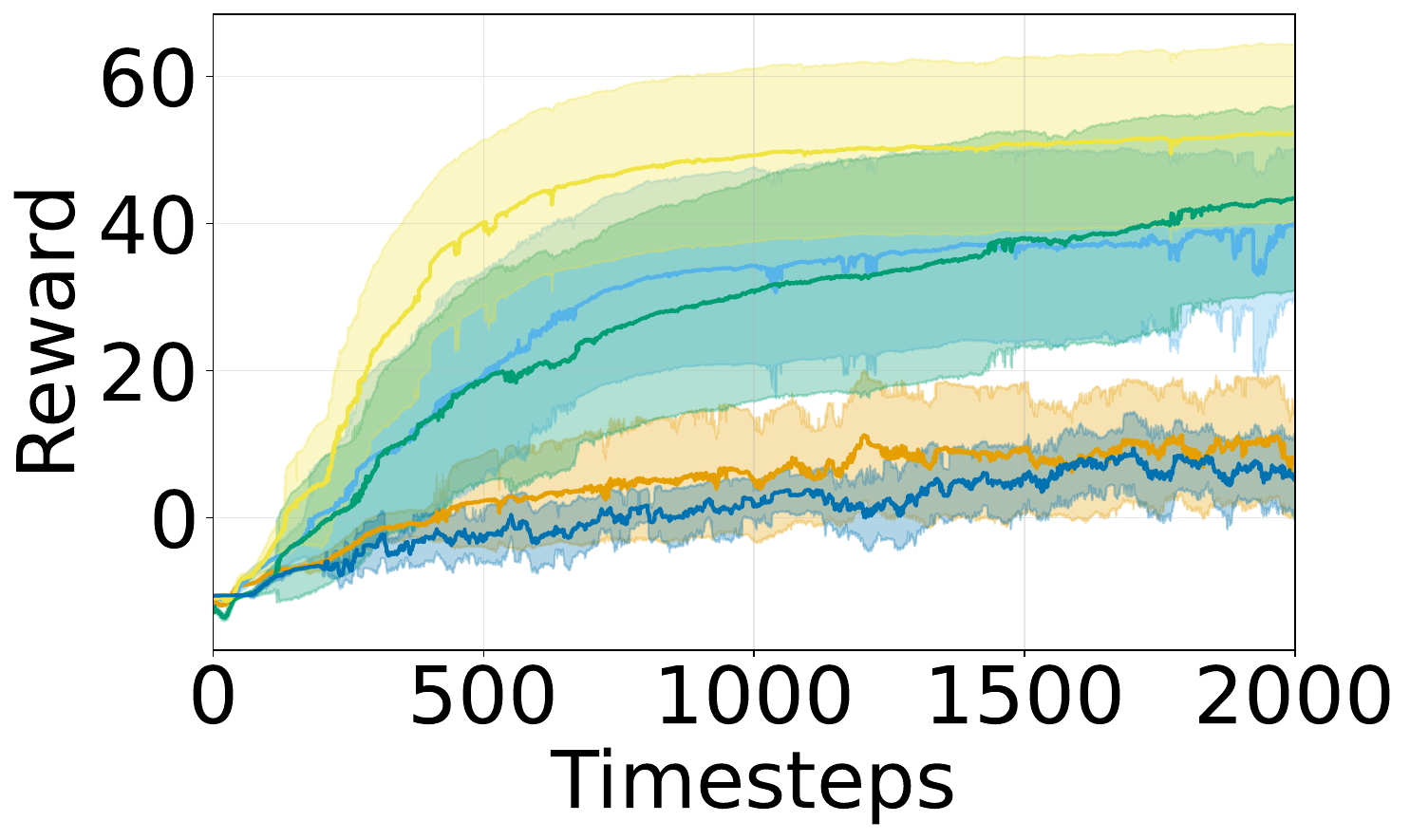}
 }
 \begin{subfigure}[b]{0.68\textwidth} 
    \centering
    \includegraphics[width=\linewidth]{images/boxplot_legend.pdf}
    \label{fig:comparison_legend}
\end{subfigure}
\caption{Episodic rewards over simulation timesteps for the conventional hexacopter and the best performing designs per task averaged and smoothed across 10 runs.}
\label{fig:episodic reward}
\vspace{-20pt}
\end{figure}

\subsubsection{Learning Dynamics }
\label{sec:analysis_of_learning_dynamics}

\begin{figure}[hbt!]
\centering

\begin{minipage}[b]{0.48\textwidth}
    \centering
    \textbf{Max Reward}
    
    \begin{minipage}[b]{0.48\textwidth}
        \centering
        \includegraphics[width=\linewidth]{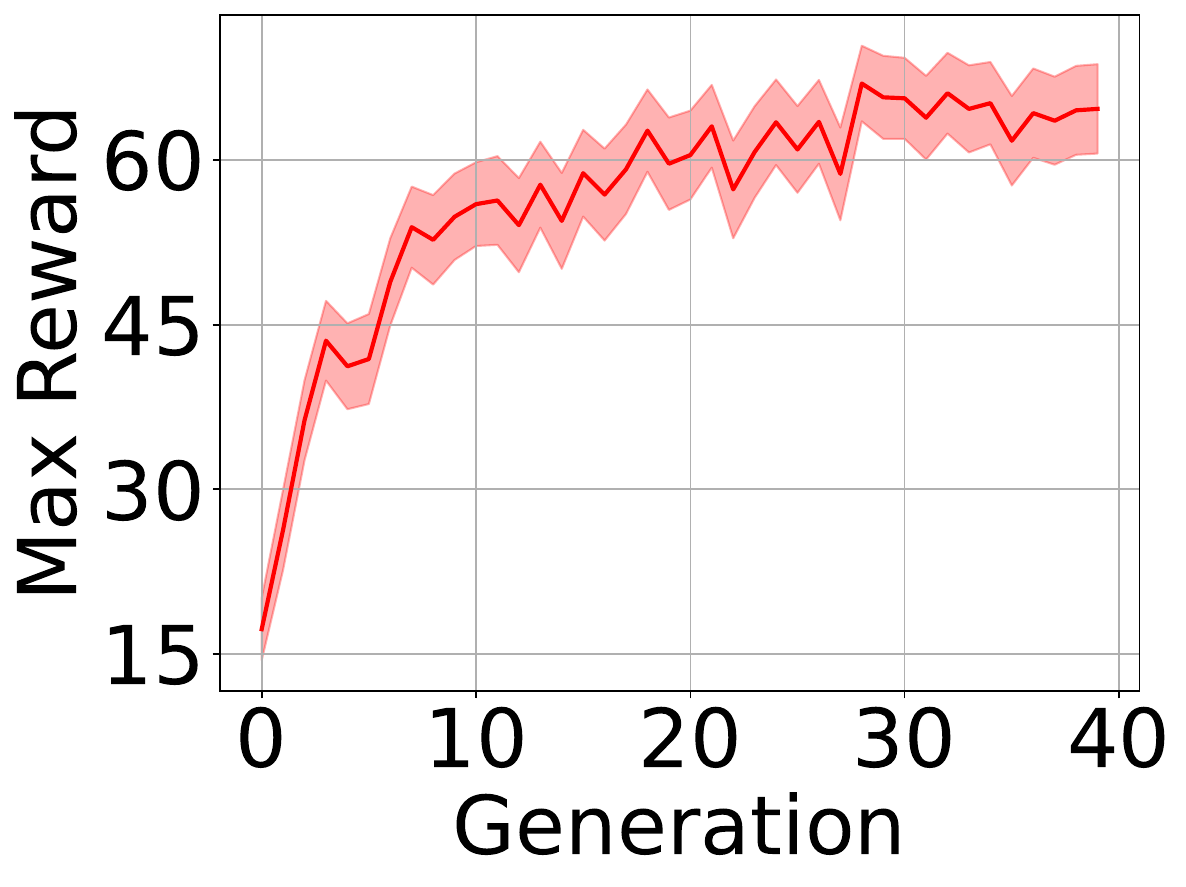}
        \textbf{(a)} Circle
    \end{minipage}
    \begin{minipage}[b]{0.48\textwidth}
        \centering
        \includegraphics[width=\linewidth]{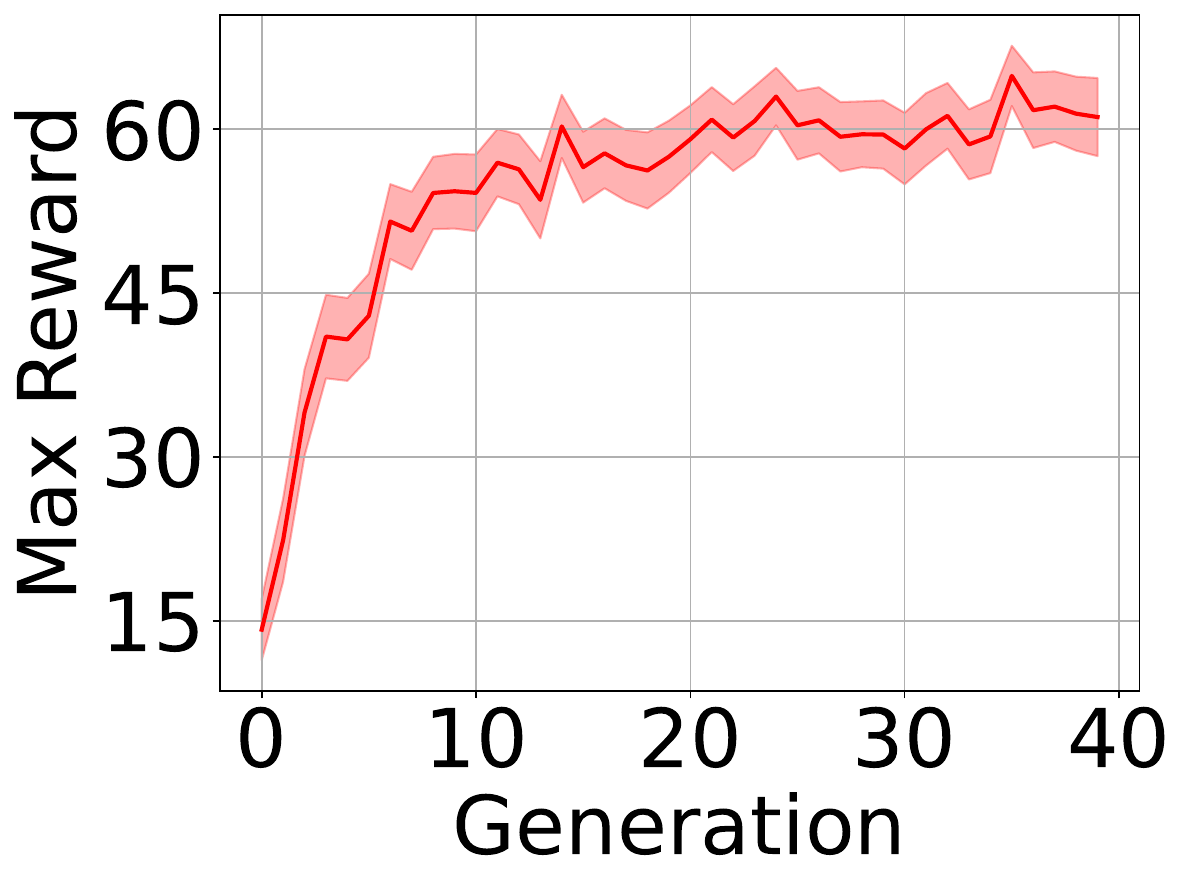}
        \textbf{(b)} Slalom
    \end{minipage}
    
    \vspace{5pt}
    \begin{minipage}[b]{0.48\textwidth}
        \centering
        \includegraphics[width=\linewidth]{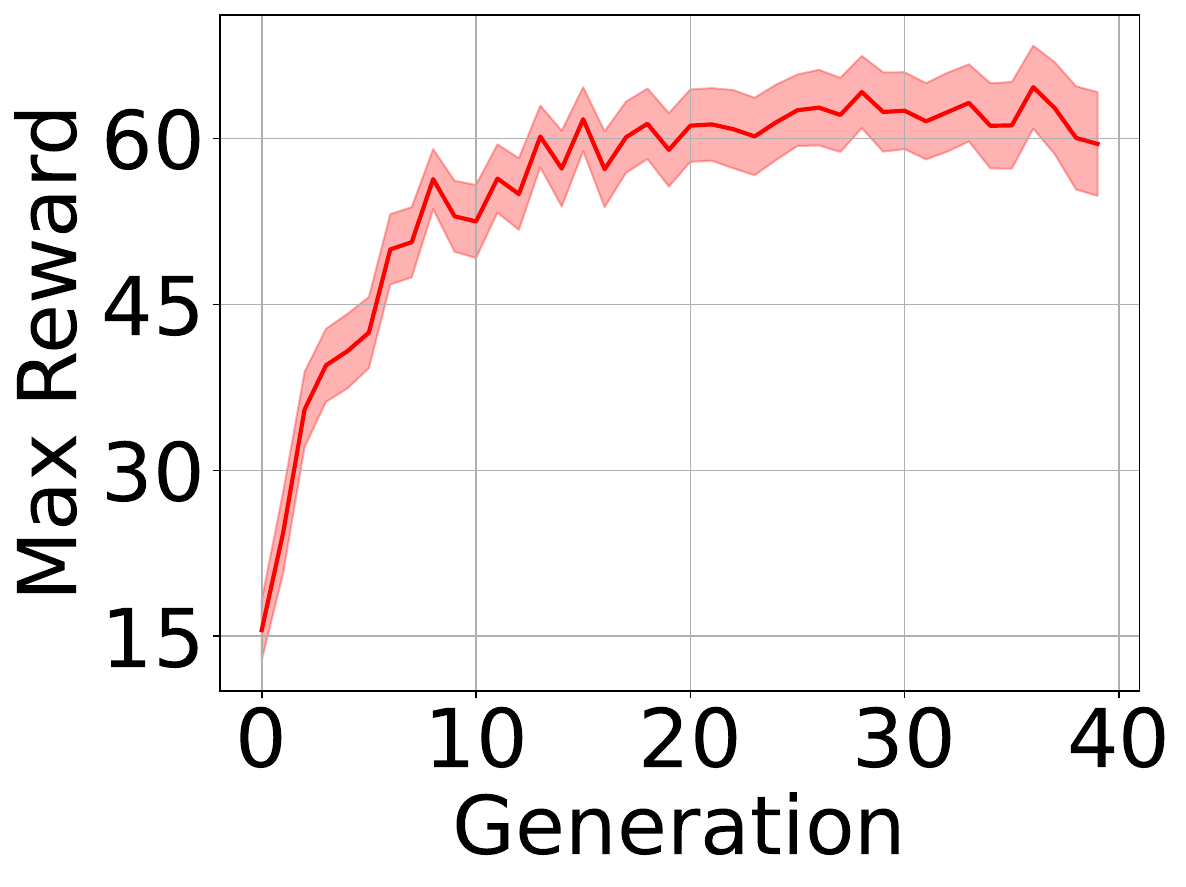}
        \textbf{(c)} Shuttlerun
    \end{minipage}
    \begin{minipage}[b]{0.48\textwidth}
        \centering
        \includegraphics[width=\linewidth]{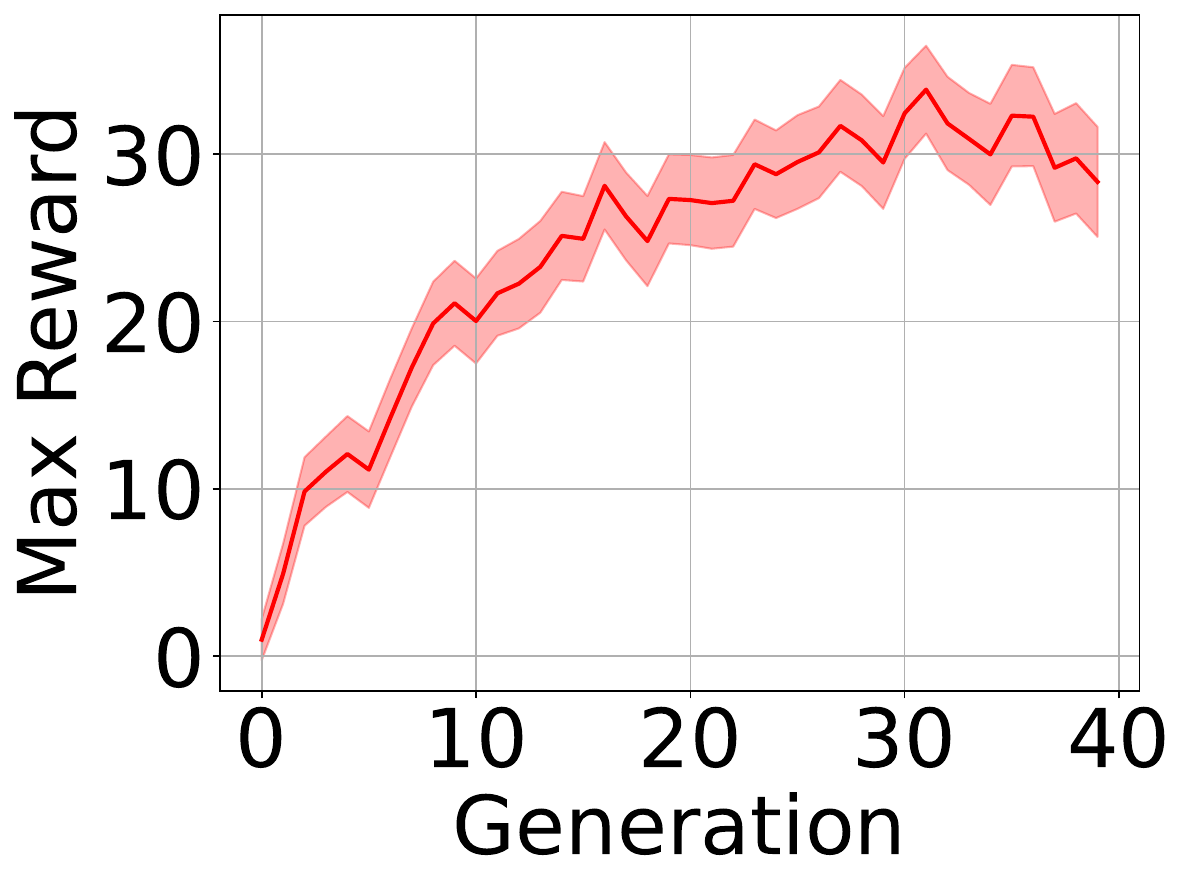}
        \textbf{(d)} Figure8
    \end{minipage}
\end{minipage}
\hfill
\begin{minipage}[b]{0.48\textwidth}
    \centering
    \textbf{Learning Speed}
    
    \begin{minipage}[b]{0.48\textwidth}
        \centering
        \includegraphics[width=\linewidth]{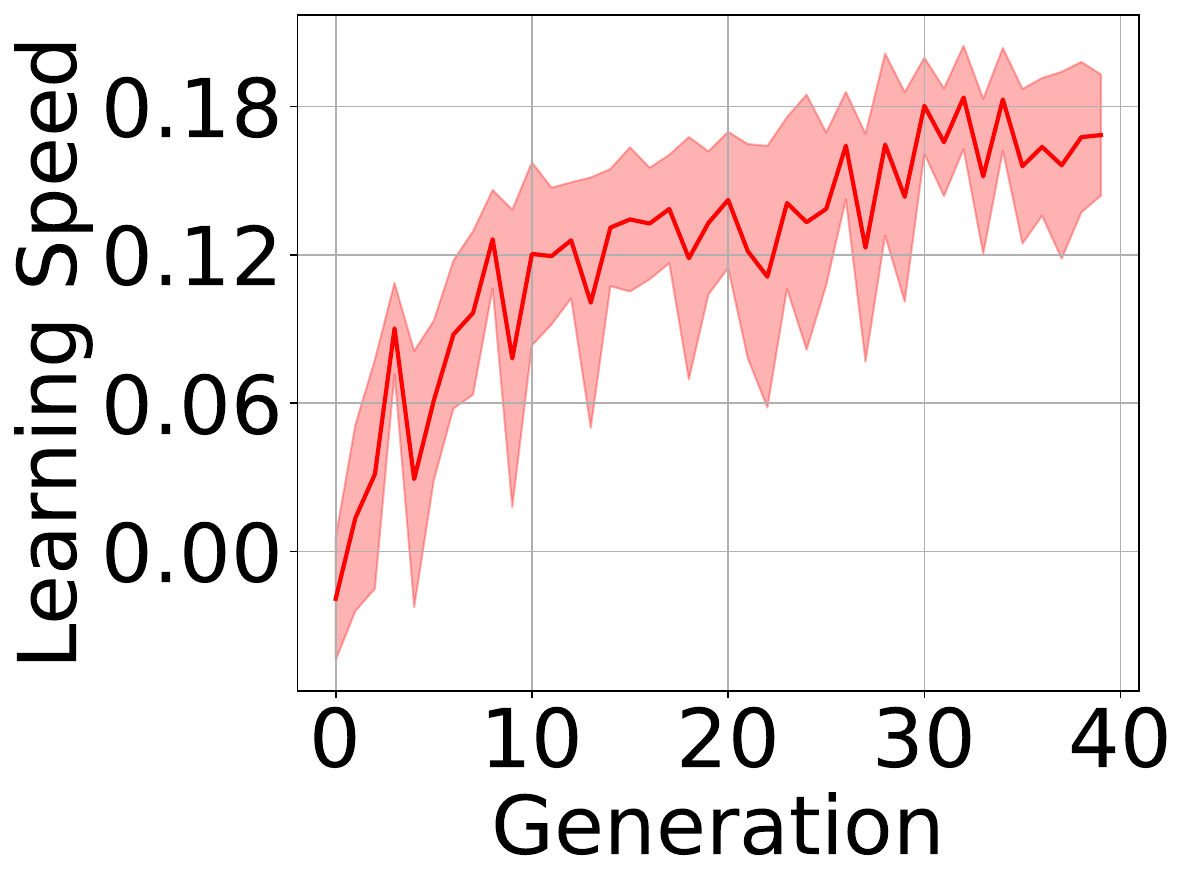}
        \textbf{(e)} Circle
    \end{minipage}
    \begin{minipage}[b]{0.48\textwidth}
        \centering
        \includegraphics[width=\linewidth]{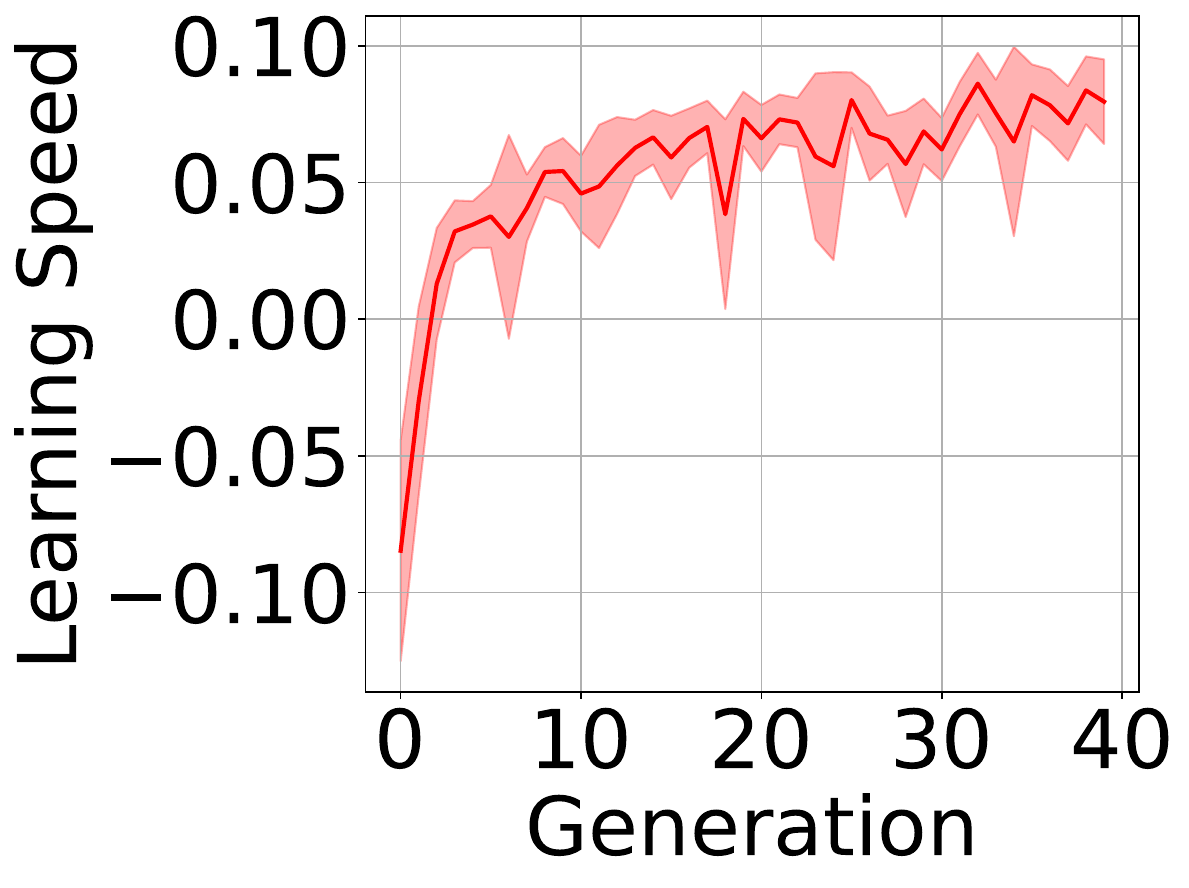}
        \textbf{(f)} Slalom
    \end{minipage}
    
    \vspace{5pt}
    \begin{minipage}[b]{0.48\textwidth}
        \centering
        \includegraphics[width=\linewidth]{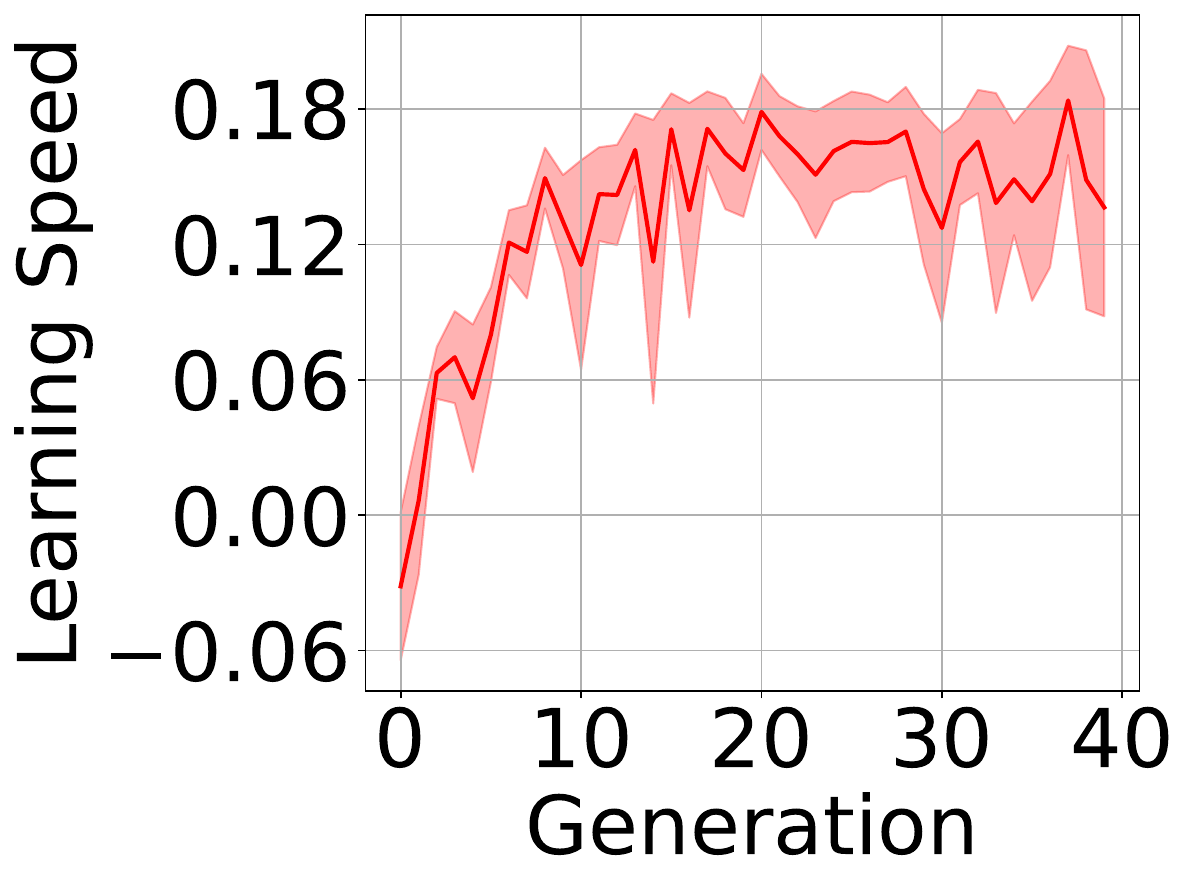}
        \textbf{(g)} Shuttlerun
    \end{minipage}
    \begin{minipage}[b]{0.48\textwidth}
        \centering
        \includegraphics[width=\linewidth]{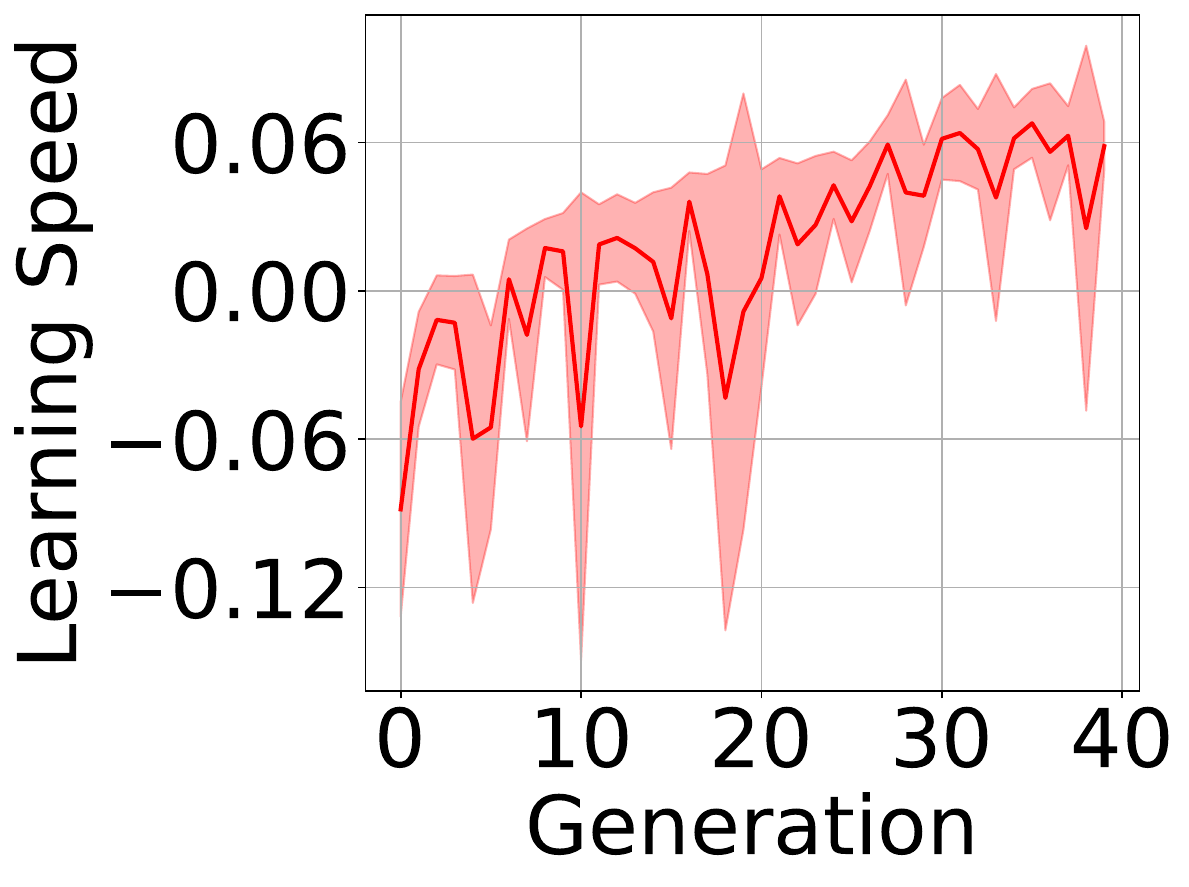}
        \textbf{(h)} Figure8
    \end{minipage}
\end{minipage}

\vspace{10pt}

\begin{minipage}[b]{0.48\textwidth}
    \centering
    \textbf{Stability}
    
    \begin{minipage}[b]{0.48\textwidth}
        \centering
        \includegraphics[width=\linewidth]{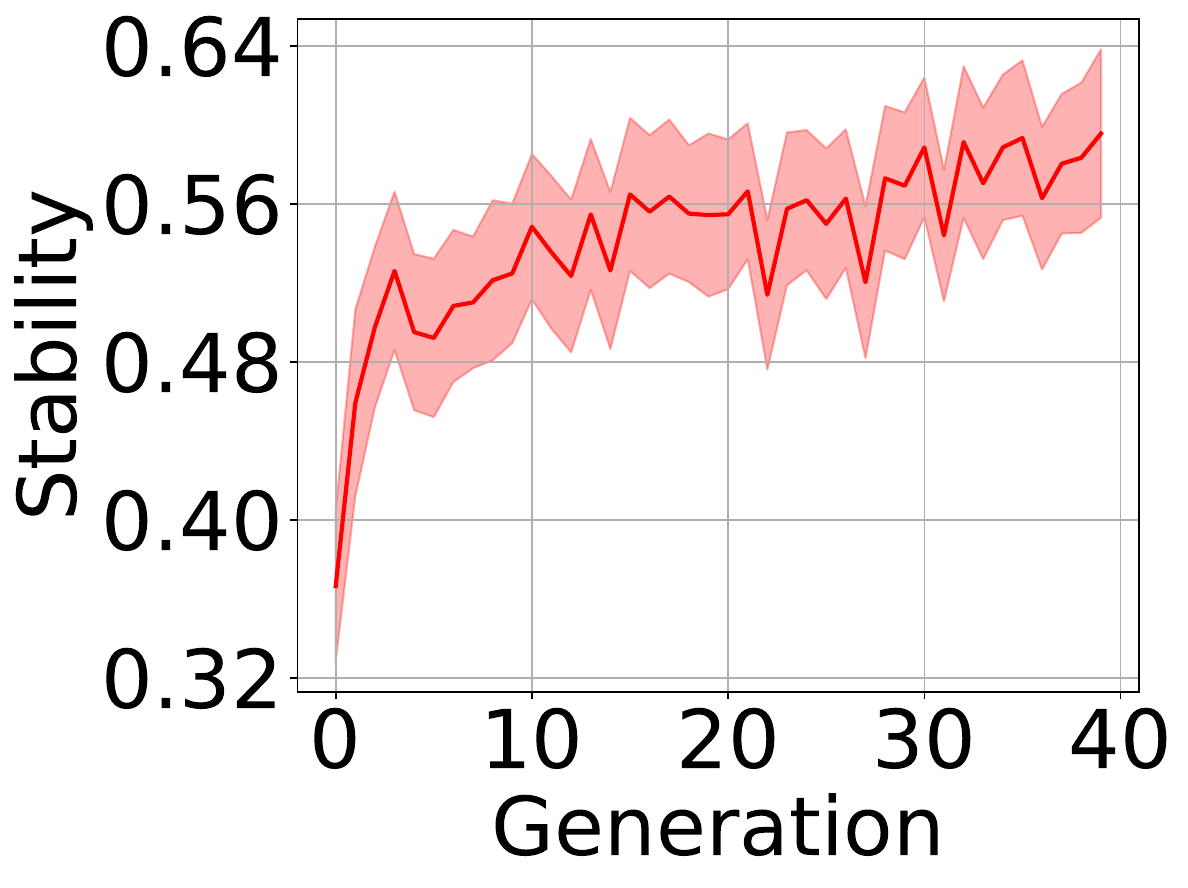}
        \textbf{(i)} Circle
    \end{minipage}
    \begin{minipage}[b]{0.48\textwidth}
        \centering
        \includegraphics[width=\linewidth]{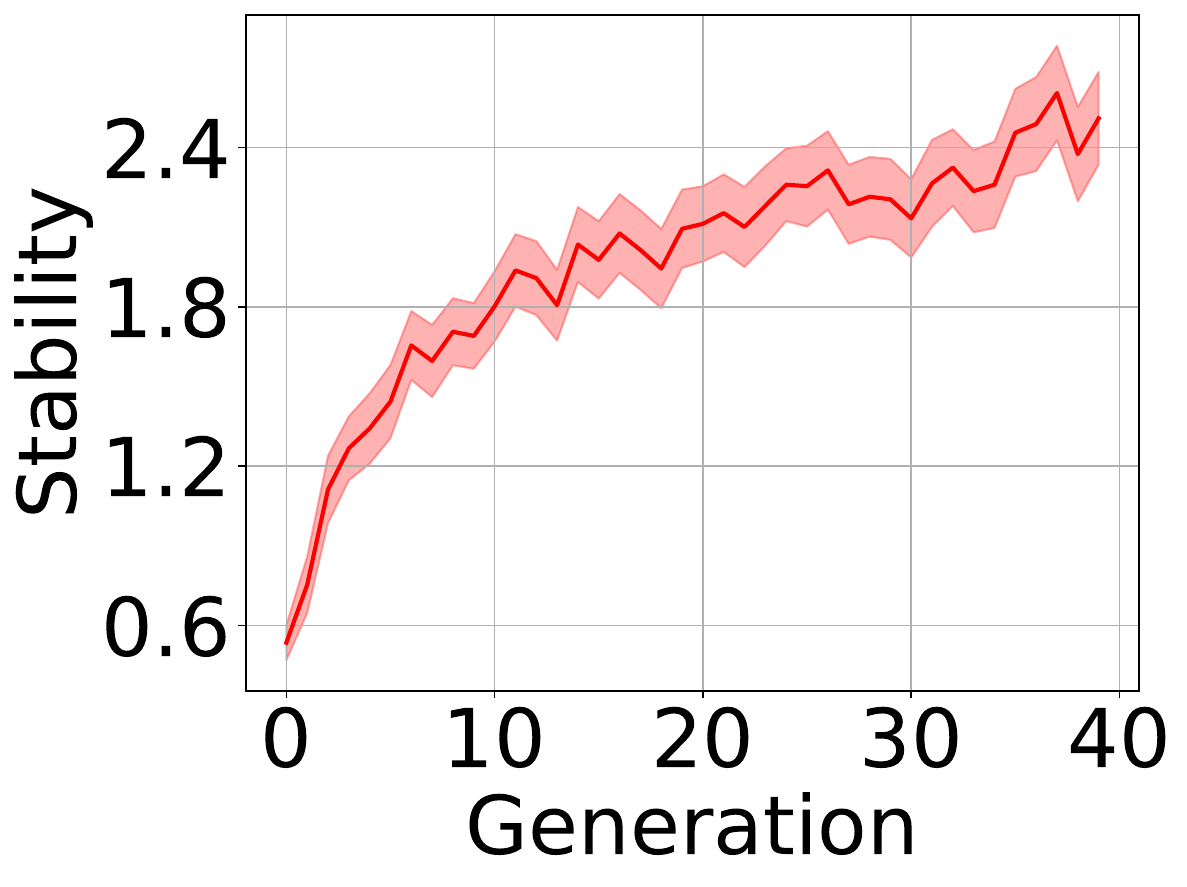}
        \textbf{(j)} Slalom
    \end{minipage}
    
    \vspace{5pt}
    \begin{minipage}[b]{0.48\textwidth}
        \centering
        \includegraphics[width=\linewidth]{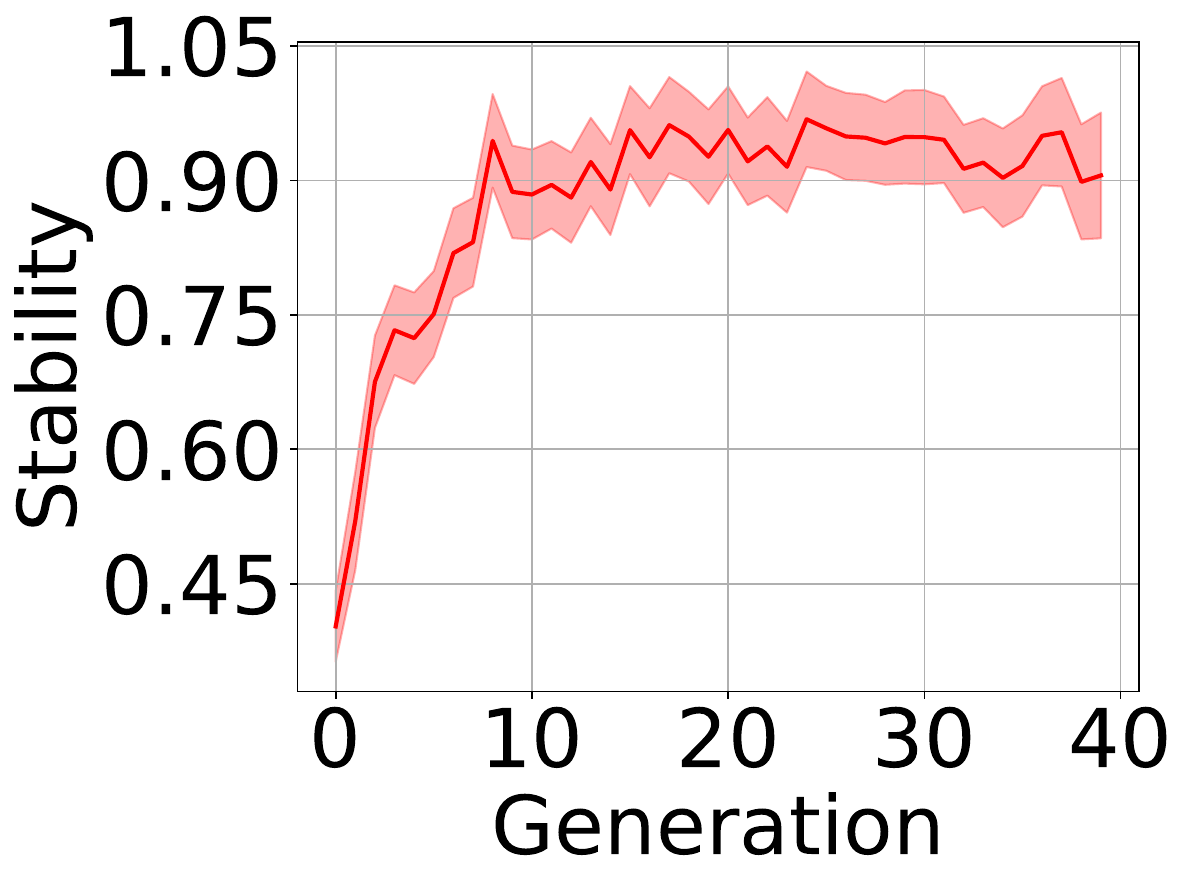}
        \textbf{(k)} Shuttlerun
    \end{minipage}
    \begin{minipage}[b]{0.48\textwidth}
        \centering
        \includegraphics[width=\linewidth]{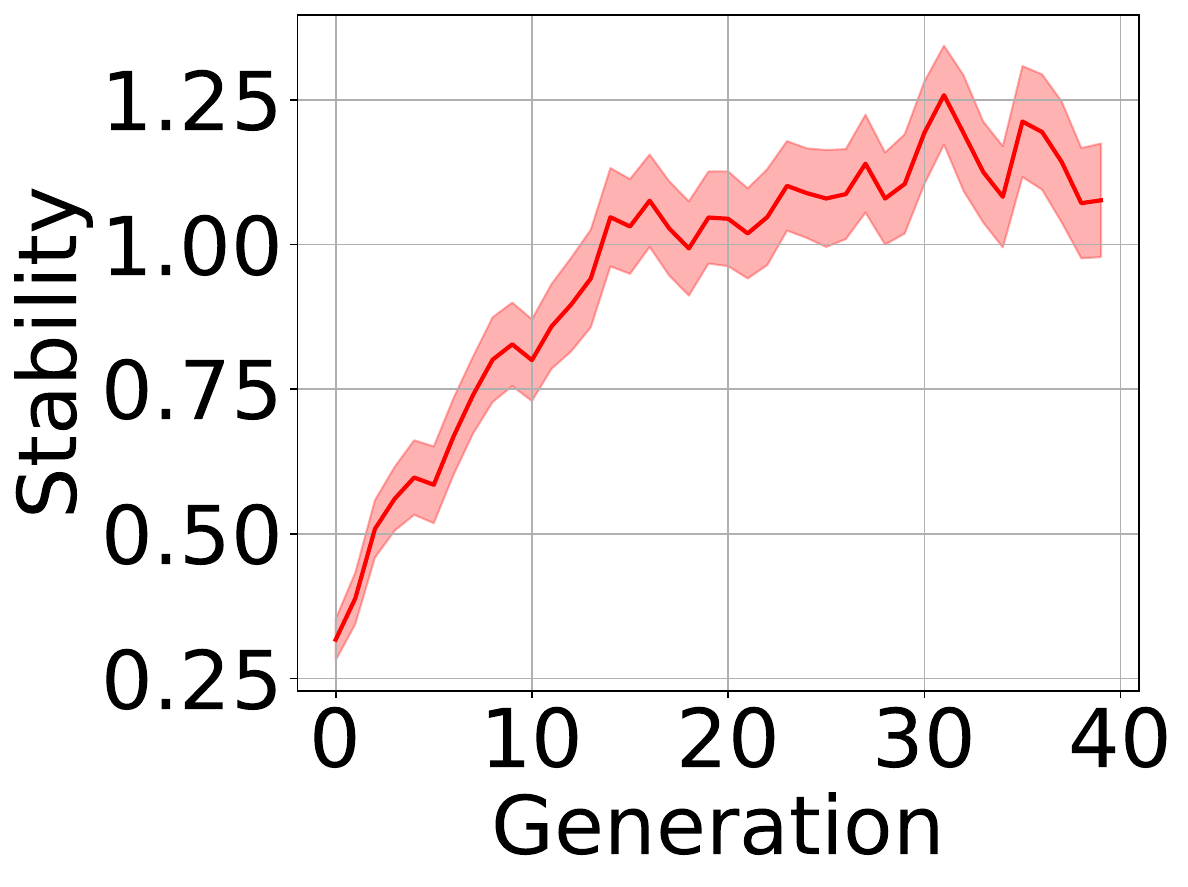}
        \textbf{(l)} Figure8
    \end{minipage}
\end{minipage}
\hfill
\begin{minipage}[b]{0.48\textwidth}
    \centering
    \textbf{Time of Fastest Learning Progress}
    
    \begin{minipage}[b]{0.48\textwidth}
        \centering
        \includegraphics[width=\linewidth]{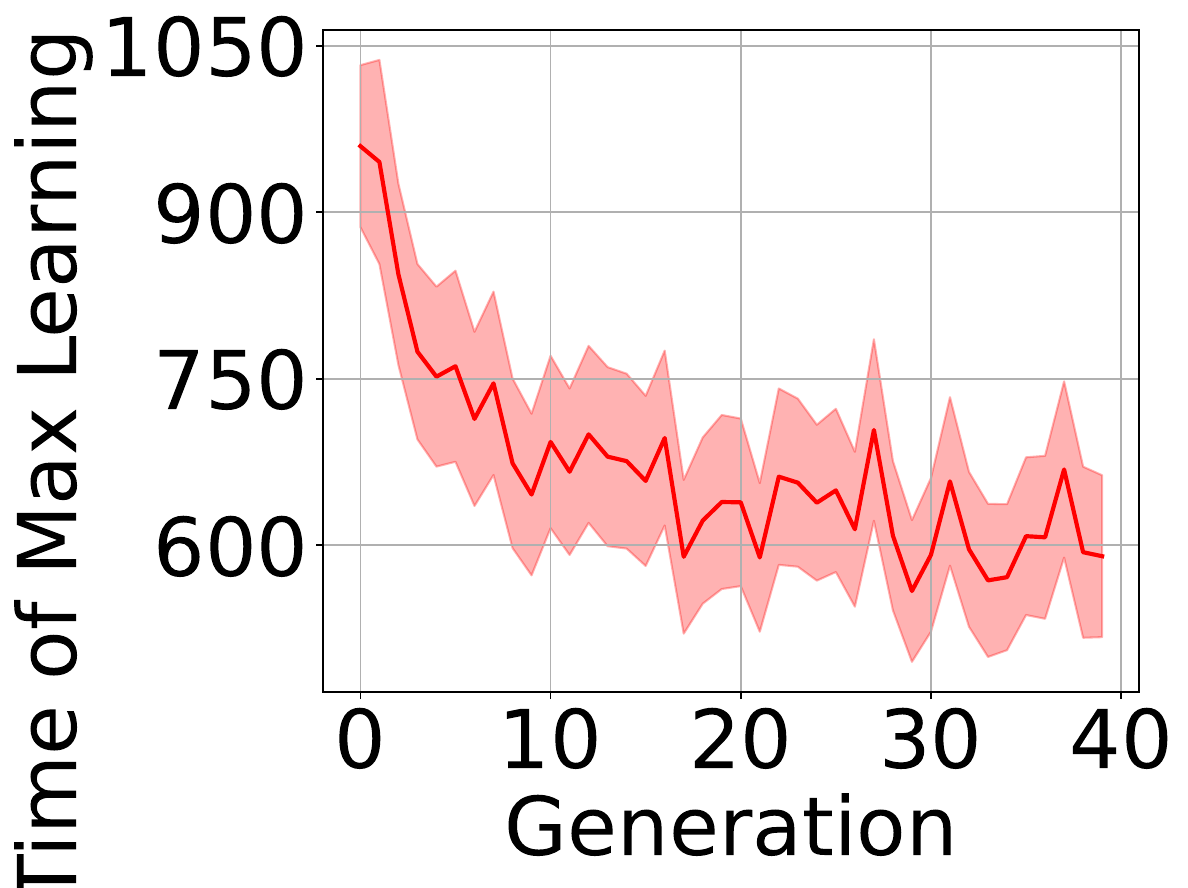}
        \textbf{(m)} Circle
    \end{minipage}
    \begin{minipage}[b]{0.48\textwidth}
        \centering
        \includegraphics[width=\linewidth]{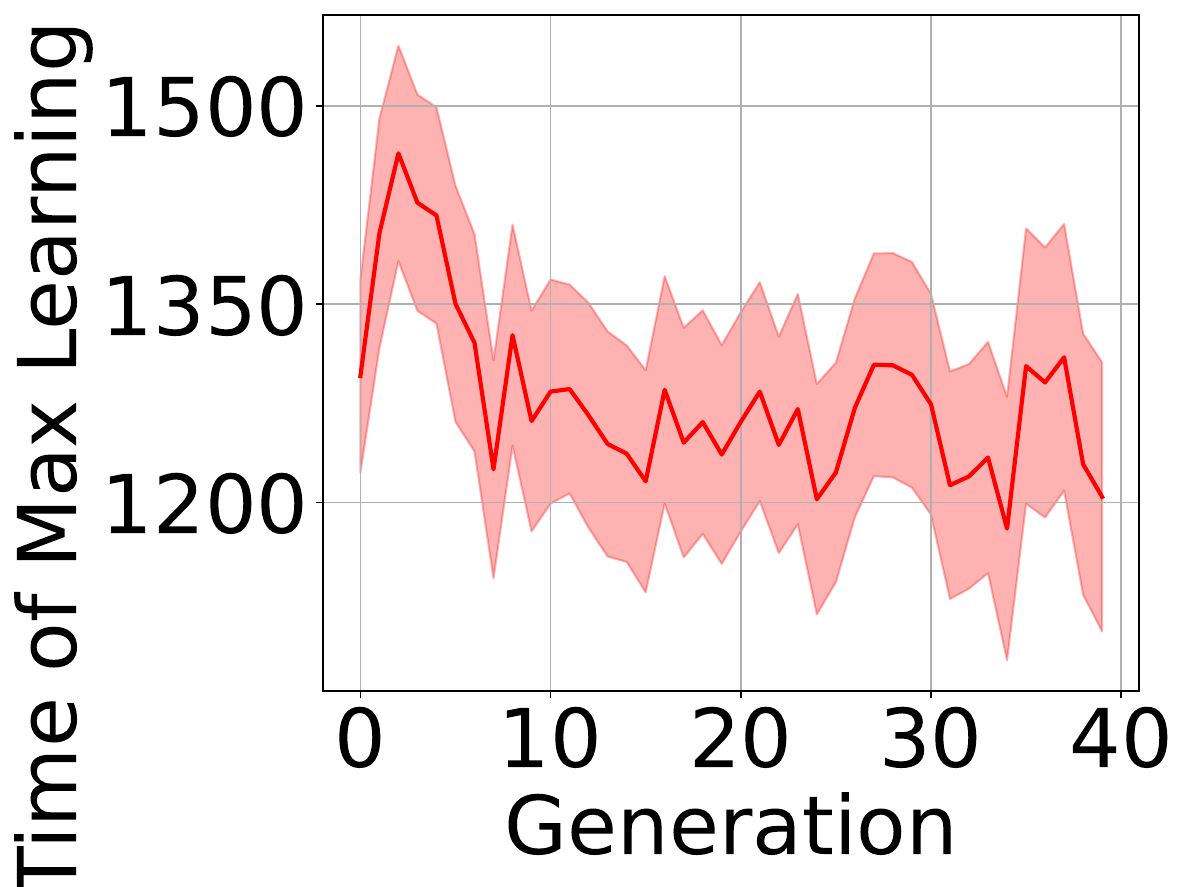}
        \textbf{(n)} Slalom
    \end{minipage}
    
    \vspace{5pt}
    \begin{minipage}[b]{0.48\textwidth}
        \centering
        \includegraphics[width=\linewidth]{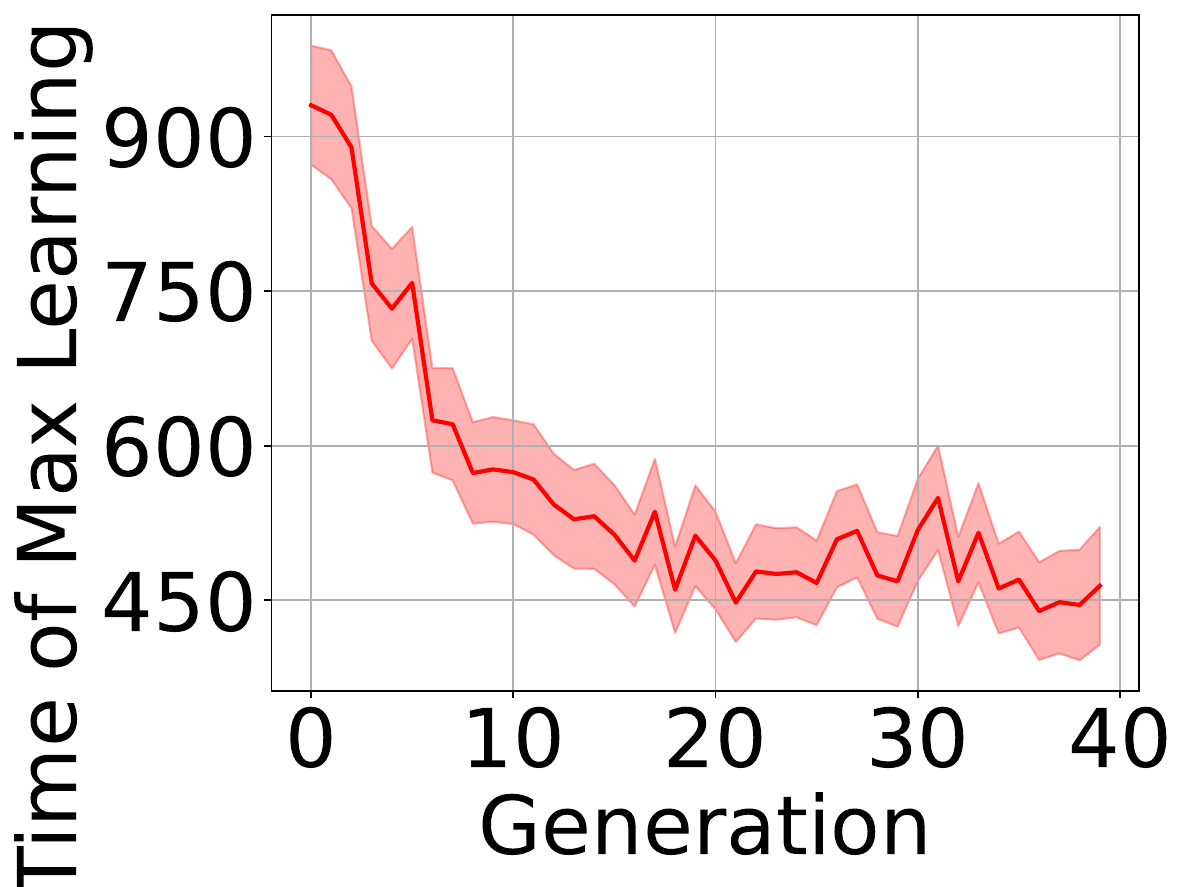}
        \textbf{(o)} Shuttlerun
    \end{minipage}
    \begin{minipage}[b]{0.48\textwidth}
        \centering
        \includegraphics[width=\linewidth]{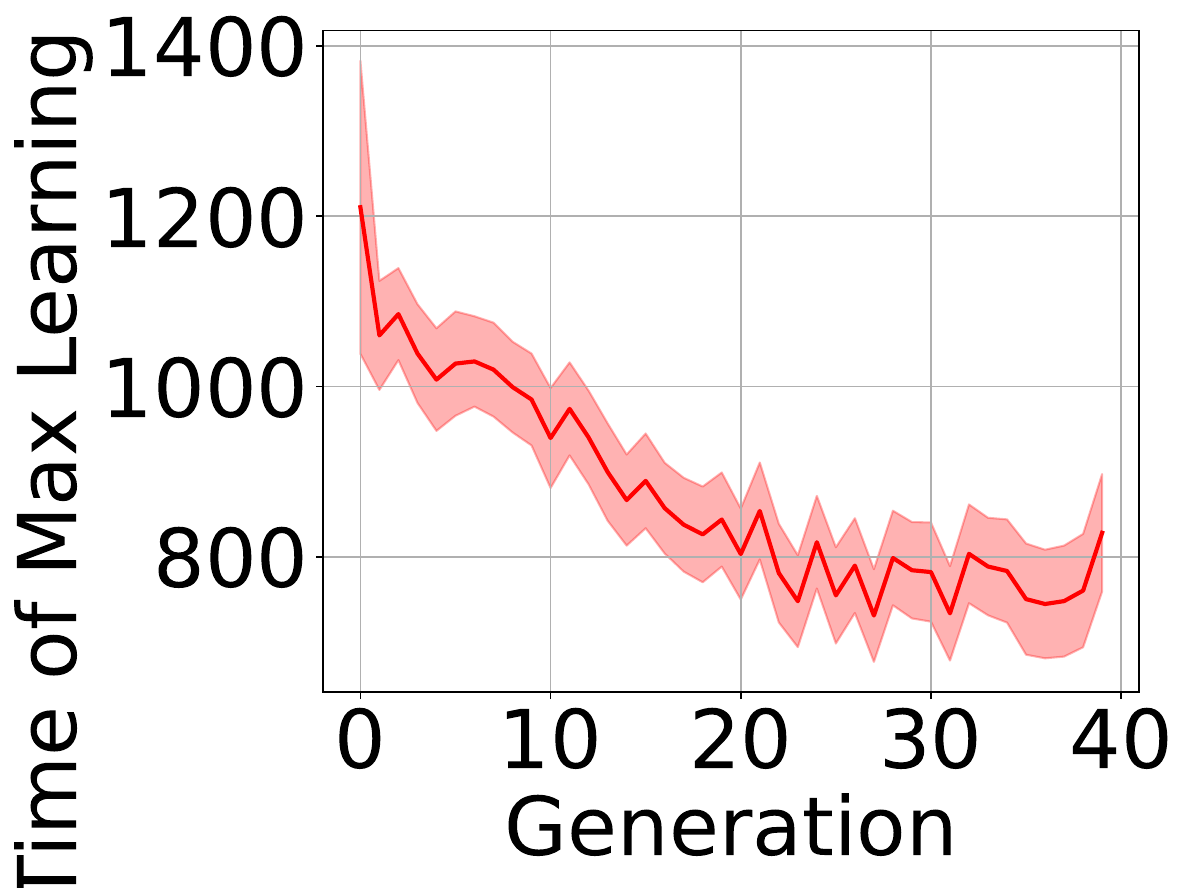}
        \textbf{(p)} Figure8
    \end{minipage}
\end{minipage}

\caption{Changes in learning efficacy and learning efficiency over generations averaged over 10 runs. Shaded regions indicate the 95\% confidence interval.}
\label{fig:learning_metrics_grid}
\vspace{-18pt}
\end{figure}

In this section, we examine the learning behavior. To this end, we first compare the learning dynamics of each hexacopter design for each task. Figure \ref{fig:episodic reward} highlights the substantial performance gains of the evolved drone morphologies compared to the conventional hexacopter. Notably, the learning curves exhibit a sharper trajectory, characterized by rapid discovery of effective behaviors followed by stable convergence. 

Figure \ref{fig:learning_metrics_grid} further illustrates the progression of learning throughout the evolutionary process. The consistently low standard deviation across runs indicates a clear and repeatable trend: evolved individuals are becoming increasingly proficient learners. Specifically, learning occurs more rapidly (figure \ref{fig:learning_metrics_grid} e-h ), and higher maximum rewards are achieved (figure \ref{fig:learning_metrics_grid} a-d). However, this improvement comes with a trade-off. On the one hand, learning is becoming less stable with increasing levels of volatility (figure \ref{fig:learning_metrics_grid} i-l ), suggesting that small policy changes have a growing impact on performance. On the other hand, the timing of the fastest learning progress is occurring earlier (figure \ref{fig:learning_metrics_grid} m-p ), indicating that the selected morphologies are those capable of acquiring fundamental flight behaviors more quickly.

Taken together, these findings suggest that the evolved designs demonstrate improved morphological intelligence, enabling them to learn effective controllers with reduced effort. This is in line with the phenomenon of increasing learning deltas as recently discovered by Miras et al \cite{Miras2020EvoLearn} and verified by others \cite{JLO-2022,9701596,WeiLi-2023}. However, in all those systems learning had a `warm start' since it started from the inherited controller, and the reward function used in the learning algorithm was identical to the fitness function used by evolution. The method in this paper uses a `cold start' since controllers are not inheritable (nor do we have an archive as in \cite{LeGoff2023}), and the reward function is different from the fitness function. 

Given these differences, the emergence of the same pattern, namely, that evolution tends to produce morphologies which support more efficient learning, across such divergent systems is striking. It suggests that the observed synergy between morphological evolution and lifetime learning is not merely a byproduct of specific algorithmic or environmental choices. Rather, it points to a deeper underlying phenomenon: a universal principle whereby evolution discovers structural priors that facilitate learning during an individual’s lifetime. This convergence across systems with different embodiments, learning paradigms, and reward structures indicates a fundamental aspect of embodied intelligence, capturing something intrinsic about the co-adaptation of body and brain.

\vspace{-1pt}
\section{Discussion and Conclusions}

Our first objective, to test whether evolution can indeed discover
unconventional morphologies that outperform traditional designs, has been clearly achieved. On all four tasks that make up our test suite, the best evolved morphologies had unconventional designs and greatly outperformed the standard hexacopter when paired with task specific learned controllers. These findings provide evidence that evolution in combination with reinforcement learning can unlock high performance solutions beyond the reach of human design intuition.

Furthermore, contrary to expectations that specialization would lead to narrow performance, evolved designs consistently outperformed the standard hexacopter across multiple tasks beyond their optimization targets. While certain designs do result in behavioral tradeoffs, other evolved morphologies exhibit strong cross task adaptability. These findings suggest that morphological evolution naturally can discover design features that confer broad performance advantages.

For our second objective, to gain insight into the interaction of morphological evolution and learning, we have introduced new metrics to analyze this interaction. These metrics can replace the learning delta from \cite{Miras2020EvoLearn} if reinforcement learning is being used as a learner within the evolutionary algorithm. Specifically, we characterized the learning dynamics through the time of fastest learning progress, learning speed, stability, and maximum reward, enabling a principled, quantitative analysis of how morphology affects learnability. This toolbox is applicable to a wide range of applications and can support investigations in embodied AI systems that combine evolution with learning.

\subsubsection{Acknowledgments}
The research reported in this paper was supported by the European Commission Horizon  project SPEAR, EU grant number 101119774, https://www.spear-robotics.com/.
\pagebreak
\bibliographystyle{splncs04}  
\bibliography{references} 

\end{document}